\documentclass[runningheads]{llncs}

\usepackage[year=2026,ID=6316]{eccv} 

\usepackage{ulem}
\usepackage{eccvabbrv}
\usepackage{graphicx}
\usepackage{booktabs}
\usepackage[accsupp]{axessibility} 
\usepackage{tikz}
\usepackage{adjustbox}
\usepackage{multirow}
\usepackage[pagebackref,breaklinks,colorlinks,citecolor=eccvblue]{hyperref}
\usepackage{orcidlink}

\begin{document}

\title{StyleFusion360: View-Consistent Head Stylization via Adaptive Style Modulation}

\titlerunning{StyleFusion360}

\author{Furkan Güzelant \and
Arda Göktoğan \and
Tarık Kaya \and
Aysegul Dundar}

\authorrunning{F.~Guzelant et al.}

\institute{Bilkent University, Department of Computer Engineering, Ankara, Türkiye\\
\email{\{furkan.guzelant, arda.goktogan, tarik.kaya\}@bilkent.edu.tr, adundar@cs.bilkent.edu.tr}}

\maketitle

\begin{figure}[ht]
 \centering
 \includegraphics[width=\linewidth]{figures/teaser_eccv_v5.pdf}
 \caption{\textbf{Multi-view 3D head stylization by StyleFusion360.} 
 Our method generates identity-preserving and multi-view consistent stylizations across diverse artistic styles, including cartoon-like and realistic domains. StyleFusion360 successfully preserves fine attributes such as accessories, facial expressions, and head geometry, while achieving high style fidelity without requiring per-style retraining. }
\label{fig:teaser}
\end{figure}

\begin{abstract}
 3D head stylization enables expressive reimagining of human faces for creative visual experiences in digital media. Existing 3D-aware methods often require computationally intensive optimization or per-style fine-tuning, limiting flexibility and user control. To overcome these challenges, we introduce StyleFusion360, a diffusion-based framework for multi-view consistent, identity-preserving 3D head stylization from a single style reference image, without per-style training. Our approach enhances the Style Fusion Attention mechanism with a style-conditioned key modulation mechanism that aligns content and style representations for fine-grained and controllable stylization. We further provide a user-controllable slider for adjusting stylization intensity. In addition, StyleFusion360 supports local multi-edit stylization, enabling targeted edits such as modifying hair or eyes independently. Extensive experiments on FFHQ and RenderMe360 demonstrate that StyleFusion360 produces high-quality, controllable, and visually compelling stylizations, outperforming state-of-the-art GAN- and diffusion-based methods across diverse style domains.
Code is available at: \url{https://github.com/furkanguzelant/StyleFusion360}

\end{abstract}


\section{Introduction}
\label{sec:intro}

3D head stylization aims to transform realistic human heads into artistic representations while preserving identity and maintaining consistency across viewpoints. This capability plays an important role in digital character creation, animation, and immersive media, where personalized and expressive avatars are increasingly demanded. Despite recent progress in generative modeling, achieving multi-view consistent, identity-preserving, and controllable stylization remains challenging.

Most stylization approaches primarily operate in the 2D image domain~\cite{patashnik2021styleclip, yang2022pastiche, pehlivan2023styleres}, where generative models enable flexible manipulation of artistic attributes. However, these methods often struggle to disentangle structural identity from stylistic transformations, which can distort facial geometry under strong stylization. Moreover, when applied independently across frames or viewpoints, they fail to maintain view consistency.
A straightforward strategy is to first stylize the input image in 2D and then reconstruct it into 3D to enforce consistency. Yet, because most 3D reconstruction models assume photorealistic inputs, stylization introduces appearance distortions that corrupt structural cues, leading to degraded geometry, identity drift, and unstable textures.
In our experiments, we evaluate several variants of such hybrid pipelines and observe consistently inferior stylization fidelity and cross-view consistency compared to performing stylization directly within a 3D-aware generative framework.

To address this limitation, recent works explore 3D-aware GAN-based generators trained for head synthesis \cite{chan2022efficient, an2023panohead}, enabling multi-view consistent rendering through volumetric or tri-plane representations. While these approaches improve geometric consistency, their visual fidelity and texture realism remain limited compared to modern diffusion models, particularly when applied to real-image stylization. Furthermore, many GAN-based stylization pipelines \cite{Bilecen2025IdentityPreserving3DHeadStylization} require domain-specific fine-tuning or optimization for each target style, reducing flexibility and scalability.

Recent advances in diffusion models have significantly improved generative quality and enabled the emergence of 3D-aware diffusion frameworks for consistent multi-view synthesis \cite{gu2025diffportrait360}. These models provide strong structural and appearance priors, but  effectively balancing structural preservation and stylistic transformation in a multi-view consistent latent space remains an open problem.
Building upon a 3D-aware diffusion architecture, we introduce \textbf{StyleFusion360}, a diffusion-based framework for \textit{multi-view consistent 3D head stylization} from a \textit{single style reference image}, without requiring per-style retraining. The framework explicitly incorporates a style-conditioned feature fusion mechanism to align content and style representations.
By performing stylization directly within a 3D-aware diffusion representation, our approach avoids the structural inconsistencies of 2D stylization pipelines while enabling controllable stylization strength and fine-grained local multi-style editing.
Our contributions:

\begin{itemize}
    \item \textbf{Style Fusion Attention:} introduces a style-conditioned key modulation mechanism that aligns content and style representations in the diffusion latent space, enabling identity-preserving yet expressive stylization with improved structural fidelity.
   \item \textbf{Local multi-style editing:} enables fine-grained region-level stylization and supports combining multiple style references (e.g., hair, eyes, and facial attributes) within a single coherent result.
    
    \item \textbf{Controllable stylization strength:} proposes a user-driven slider for continuous adjustment of stylization intensity.
    
\item \textbf{Multi-view diffusion training:} uses synthetic 3D GAN multi-view data to learn cross-view consistency, yet generalizes effectively to real images at inference, producing higher-quality stylizations without being constrained by the visual fidelity of the synthetic training data.

    \item \textbf{Arbitrary style generalization:} transfers unseen styles at inference time without per-style optimization or retraining, demonstrating strong generalization across diverse artistic domains.
\end{itemize}

Comprehensive experiments on FFHQ and RenderMe360 demonstrate that StyleFusion360 achieves superior style fidelity, high-quality, and multi-view consistency compared to state-of-the-art GAN- and diffusion-based stylization methods, while also faithfully preserving fine details and accessories such as hats and glasses as shown in Fig. \ref{fig:teaser}.
Our results demonstrate the effectiveness of combining 3D-aware diffusion with controllable style adaptation for high-quality 3D head stylization.
\section{Related Work}

\noindent \textbf{3D-Aware Generators.} 
Generative Adversarial Networks (GANs) have made significant progress in generating 3D-aware and view-consistent images using both explicit and implicit representations~\cite{dundar2023fine, pavllo2020convolutional, chan2021pi, or2022stylesdf, chan2022efficient}. Notable works on human face generation include PanoHead~\cite{an2023panohead} and SphereHead~\cite{li2024spherehead}. While these methods achieved impressive results, back-view synthesis remains challenging, and they often struggle to represent accessories or complex hairstyles accurately.

Diffusion models have recently demonstrated superior performance for 3D-aware generation. DiffPortrait3D~\cite{gu2024diffportrait3d} integrates appearance and camera-view attentions into pre-trained U-Nets for novel view synthesis, achieving strong results for near-frontal portraits but sometimes producing local inconsistencies. DiffPortrait360~\cite{gu2025diffportrait360} extends this framework to enable full 360-degree portrait generation, maintaining identity consistency across poses via a customized ControlNet for back-of-head detail generation and a dual-appearance module that enforces global front–back consistency. We build upon DiffPortrait360 to enable 360$^{\circ}$ head stylization.

\noindent \textbf{Style Transferring in Image Generation.} Early research on generative stylization primarily focused on the 2D image domain, where neural style transfer \cite{jing2019neural} and generative models \cite{li2018closed, deng2022stytr2} enable flexible manipulation of artistic attributes.
More recent works employ generative models such as StyleGAN ~\cite{kerras2018stylegan, liu2021FastGAN, ojha2021few-shot-gan,yang2022pastiche} and diffusion-based frameworks \cite{wang2025omnistyle, chung2024styleid, brooks2023ip2p, wang2025domain} to perform reference-based or text-guided stylization, producing visually compelling results across diverse artistic domains. While these methods demonstrate strong visual quality in single images, they lack geometric awareness and therefore cannot maintain consistency across different viewpoints.
Applying 2D stylization independently across views generally leads to inconsistent results. Several works therefore combine image stylization with subsequent 3D reconstruction or novel-view synthesis. However, these pipelines remain challenging due to the mismatch between stylized appearances and the assumptions of downstream 3D models. Motivated by these limitations, recent works have explored 3D-aware generative approaches that jointly model appearance and geometry \cite{lei2024diffusiongan3d, song2024DiffusionGuidedDomainAdaptation, Bilecen2025IdentityPreserving3DHeadStylization, oztas20253dstylizationlrm}.

\section{Method}

We present StyleFusion360, a diffusion-based framework for multi-view consistent 3D head stylization from a single style reference image. Our approach introduces a style-conditioned feature fusion mechanism that enables controllable and identity-preserving stylization within a 3D-aware diffusion representation.
Our framework is built on top of a 3D-aware diffusion backbone for portrait generation, which provides strong structural priors for multi-view consistency. Rather than modifying the underlying geometry representation, we focus on enabling effective style integration while preserving structural identity.

Specifically, \cref{preliminaries} briefly reviews the 3D-aware diffusion backbone used in our framework.  
\cref{style_fusion_attention} introduces our proposed \textit{Style Fusion Attention}, which injects style information into the diffusion process through a style-conditioned key modulation mechanism.  
\cref{fine-tuning} describes our multi-view diffusion training strategy using synthetic data.  
\cref{local_stylization} presents our region-aware stylization strategy that enables fine-grained local editing and multi-style fusion.  
\ref{key_scaling} presents our controllable stylization mechanism for adjusting style strength at inference. Additional implementation details are provided in the supplementary material.


\subsection{Preliminaries}
\label{preliminaries}

\noindent\textbf{3D-aware Diffusion Backbone.}
Our framework builds upon a 3D-aware latent diffusion architecture designed for multi-view portrait synthesis \cite{gu2025diffportrait360}. 
Given a reference portrait and a target camera pose, the model generates view-consistent head images across a full 360° range by leveraging pose conditioning and cross-view feature interaction within the diffusion process.
The backbone adopts a frozen latent diffusion model (LDM) \cite{rombach2022ldm} as the base generator and incorporates additional trainable modules for appearance conditioning and multi-view consistency. 
To improve structural stability across viewpoints, the model is trained using multi-view supervision together with 3D-aware noise priors derived from synthetic data \cite{an2023panohead}.
While this architecture provides strong geometric and identity priors for novel-view synthesis, integrating expressive and controllable stylization into a multi-view consistent diffusion representation remains an open problem. 
We tackle this challenge through a style-conditioned feature fusion mechanism that enables identity-preserving stylization directly in the diffusion latent space.

\subsection{Style Fusion Attention}
\label{style_fusion_attention}

Generating multi-view consistent stylized head images requires balancing 
structural identity from a content image \(\mathrm{I}_{\text{content}}\) 
with stylistic attributes from a reference image \(\mathrm{I}_{\text{style}}\). 
While the 3D-aware diffusion backbone provides strong geometric priors, 
it does not explicitly disentangle content and style representations, 
which is crucial for identity-preserving stylization.

\begin{figure*}[t]
  \centering
  \includegraphics[width=\linewidth]{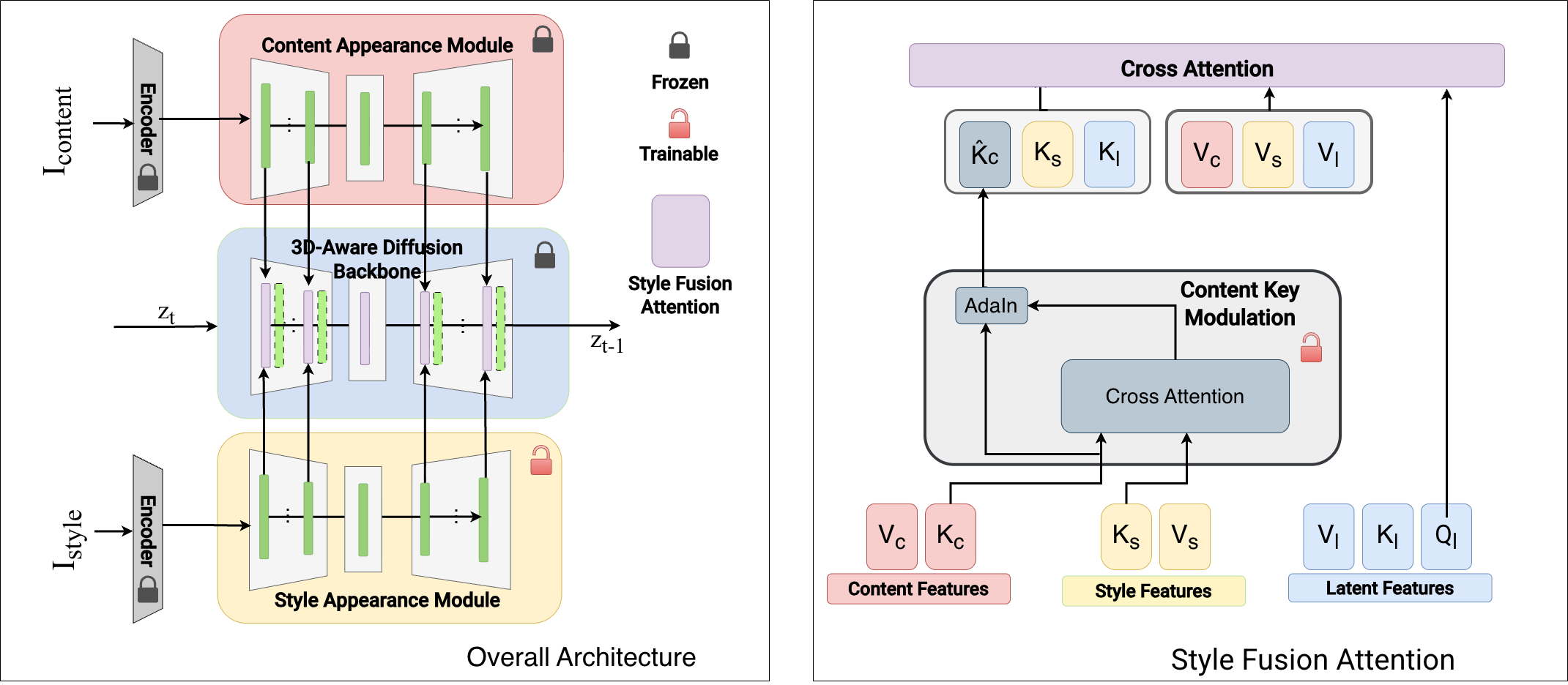}
  \caption{
  \textbf{An overview of our proposed framework.}  
\textbf{(a) Architecture Overview:} The model takes a content image \(\mathrm{I}_{\text{content}}\), a style reference \(\mathrm{I}_{\text{style}}\), and a target camera pose \(\mathrm{I}_{\text{cam}}\). Separate Content and Style Appearance Modules extract identity and style features, which are fused in the main U-Net through \textbf{Style Fusion Attention} to produce multi-view consistent stylized outputs.  
\textbf{(b) Style Fusion Attention Mechanism:} Latent features from the main U-Net act as queries, attending to a combination of content and style keys and values. Content keys are modulated via content key modulation, enabling balanced stylization while preserving identity.
        }
\label{fig:overall}
  \label{fig:overall}
\end{figure*}

The \textbf{Content Appearance Module}, adopted from the 3D-aware diffusion backbone, 
encodes structural and identity-related features from the content image. 
To preserve geometry and identity consistency, this module is kept frozen during training. 
In contrast, stylization requires capturing attributes such as color, texture, 
and artistic patterns. We therefore introduce a dedicated 
\textbf{Style Appearance Module}, initialized from the content encoder 
but trained to extract style-specific representations from 
\(\mathrm{I}_{\text{style}}\). This disentangled design as shown in Fig. \ref{fig:overall} ensures that 
content and style features are encoded separately.

A key challenge arises from the frozen content pathway. 
Global style alignment methods such as StyleAligned~\cite{hertz2024stylealigned} align global feature statistics 
between content and style representations, but apply uniform modulation 
across spatial locations. Such global conditioning lacks spatial selectivity 
and is insufficient for semantically structured regions such as hair, eyes, 
or skin. This limitation is particularly critical in our setting, where the content pathway is frozen and may otherwise dominate the fusion process. We intentionally keep the content pathway frozen to avoid degrading the generation quality when training with synthetic data.

To address this issue, we introduce a two-stage attention mechanism 
that explicitly modulates the content key before shared attention as shown in Fig. \ref{fig:overall}.

\noindent
\textbf{Feature Projection.}
Given latent diffusion features $\mathbf{F}_l$, content features $\mathbf{F}_c$, 
and style features $\mathbf{F}_s$, we compute their corresponding query, key, 
and value representations using standard learnable linear projections 
as in the attention mechanism:
\begin{equation}
(\mathbf{Q}_l, \mathbf{K}_l, \mathbf{V}_l), \quad
(\mathbf{K}_c, \mathbf{V}_c), \quad
(\mathbf{K}_s, \mathbf{V}_s).
\end{equation}

\vspace{0.5em}
\noindent
\textbf{Stage 1: Content Key Modulation.} 
We first introduce a content key modulation stage to adapt the spatial structure of the content features according to the style reference.
A naive approach would process the style embedding with a global mapping (e.g., an MLP) and apply a uniform modulation to the content representation. However, such global transformations ignore spatial correspondences between the content and style images. In practice, stylistic characteristics often vary across regions (e.g., shading, facial structure, or material properties), requiring region-dependent modulation.

Furthermore, the content and style images are not spatially aligned and may appear under different viewpoints. While spatially varying modulation could in principle be obtained using convolutional mappings, such approaches implicitly assume spatial correspondence between the two feature maps. In our setting, this assumption does not hold, making convolution-based normalization unreliable. Therefore, we instead employ a cross-attention mechanism to discover soft correspondences between content and style features before computing the spatial modulation.
To address this, we compute a cross-attention in the key space, allowing the model to discover soft spatial correspondences between content and style features and produce a spatially varying modulation signal.

We perform a dedicated cross-attention in the key space 
to modulate the content key using style information. 
Let $\mathbf{K}_c$ and $\mathbf{K}_s$ denote the projected 
content and style key embeddings.
These projected keys are then fed into a secondary cross-attention 
module that introduces additional learnable linear mappings 
specific to the modulation stage. In this module, 
$\mathbf{K}_c$ after a projection layer serves as the query, while transformed versions 
of $\mathbf{K}_s$ provide the key and value representations:

\begin{equation}
\mathbf{K}_s^{m}, \mathbf{V}_s^{m}
=
\text{Proj}(\mathbf{K}_s).
\end{equation}

The cross-attention is computed as

\begin{equation}
\mathbf{F}_{cs}
=
\text{Softmax}
\left(
\frac{\mathbf{K}_c^{m}(\mathbf{K}_s^{m})^{\top}}{\sqrt{d}}
\right)
\mathbf{V}_s^{m}.
\end{equation}

Finally, the resulting feature modulates the original content key using Adaptive Instance Normalization (AdaIN)~\cite{huang2017arbitrary}, which transfers style by matching the feature statistics of a content representation to those of a style representation:

\begin{equation}
\hat{\mathbf{K}}_c
=
\text{AdaIN}(\mathbf{K}_c, \mathbf{F}_{cs}).
\end{equation}

\vspace{0.5em}
\noindent
\textbf{Stage 2: Shared Cross-Attention.} In the final attention layer of the diffusion backbone, 
we perform shared attention using:

\begin{equation}
\mathbf{Q} = \mathbf{Q}_l,
\quad
\mathbf{K} = [\mathbf{K}_l, \hat{\mathbf{K}}_c, \mathbf{K}_s]^{\top},
\quad
\mathbf{V} = [\mathbf{V}_l, \mathbf{V}_c, \mathbf{V}_s]^{\top}.
\end{equation}

Notably, the style embeddings \( \mathbf{K}_s \) and 
\( \mathbf{V}_s \) are directly reused here without the 
additional modulation projection applied in Stage 1.

The final attention is computed as:

\begin{equation}
\mathbf{A}
=
\text{Softmax}
\left(
\frac{\mathbf{Q}\mathbf{K}^{\top}}{\sqrt{d}}
\right)\mathbf{V}.
\end{equation}

By explicitly modulating the content key prior to shared attention, 
our design prevents the frozen structural pathway from dominating 
the fusion process while preserving identity consistency. 
At the same time, direct style embeddings remain available in 
shared attention, enabling strong stylistic guidance. 
This two-stage formulation achieves spatially selective, 
identity-preserving stylization across multiple views.

\subsection{Multi-view Diffusion Training for Consistent Stylization}
\label{fine-tuning}

We fine-tune the \textbf{Style Appearance Module} and the \textbf{Style Fusion Attention Module}  using a GAN-generated paired multi-view dataset. Specifically, we adopt a 3D GAN~\cite{Bilecen2025IdentityPreserving3DHeadStylization} that is stylized for a target domain by fine-tuning a base panoramic head generator~\cite{an2023panohead}. Although the resulting paired images (original vs.\ stylized) inherit the limited photorealism of GAN-based synthesis, they provide consistent head geometry and well-aligned stylization across multiple views. Crucially, since the \textit{Content Appearance Module} remains frozen and was pretrained on high-quality real faces, the network learns to produce realistic and identity-preserving stylizations even when presented with real content images, effectively bypassing the limitations of the synthetic training data.

During training, stylized reference images are chosen to depict identities different from the content image, encouraging disentanglement between style and identity. The extracted content, style, and pose features are fused in the main SD-UNet via \textit{Style Fusion Attention} and the trainable \textit{View Consistency Module}, enabling multi-view consistent stylized outputs. Unlike prior GAN-based stylization approaches~\cite{Bilecen2025IdentityPreserving3DHeadStylization}, which require domain-specific training for each style, our method generalizes to arbitrary new styles at inference without additional fine-tuning or per-style optimization.

In practice, domain-specific GAN stylization for a small set of representative styles (e.g., six styles, 3 hours per style) followed by 2 hours of Style Appearance Module and Attention Module training constitutes a one-time cost. After this stage, the model can stylize any new image with a novel style in roughly 3 minutes, making the approach efficient and flexible compared to methods requiring retraining for each style.
We provide the runtime analysis of all methods in Supplementary.

\subsection{Region-Aware Local Stylization}
\label{local_stylization}

Our framework supports fine-grained, region-specific stylization (e.g., hair, eyes, or mouth) using a \textbf{style mask}. 
Given a style reference image \(\mathrm{I}_{\text{style}}\), we generate binary masks \( \mathbf{M} \in \{0,1\}^{H \times W} \) for the target regions using an off-the-shelf segmentation model \cite{ravi2024sam2}. 

These masks are resized to match the spatial resolution of the latent features and applied to modulate the style key features \( \mathbf{K}_s \) within the Style Fusion Attention layers via element-wise multiplication. 
To further control stylization strength, we employ Adaptive Instance Normalization (AdaIN), applying the AdaIN-transformed style features \( \hat{\mathbf{K}}_c \) in masked regions and retaining the original content features \( \mathbf{K}_c \) elsewhere. 

This formulation allows precise, region-level stylization while preserving identity and structural consistency in non-edited areas. 
Additionally, it naturally supports \textit{multi-style editing}, enabling different regions to be stylized with separate reference images within a single diffusion pass.

\begin{figure*}[t]
    \centering
    \includegraphics[width=\linewidth]{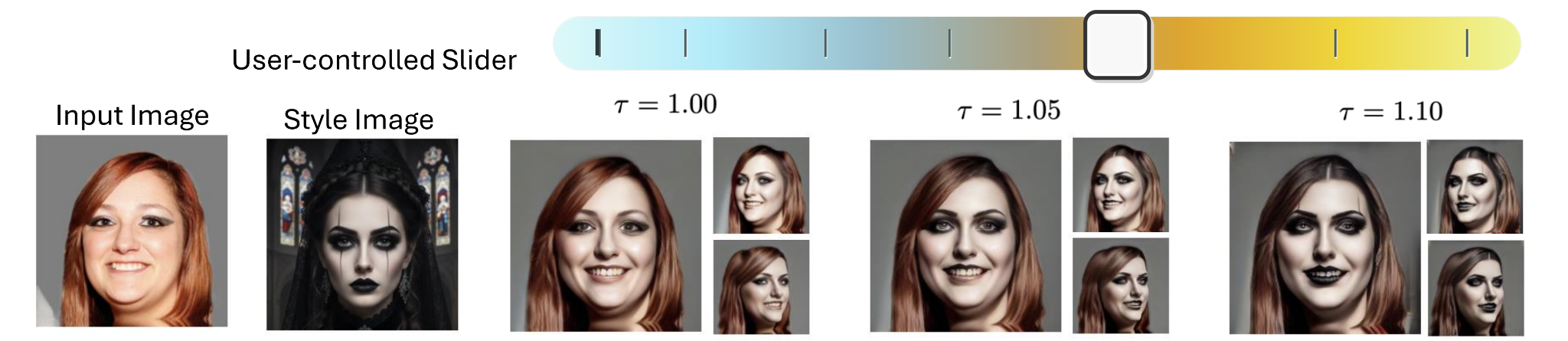}
 \caption{Controllable stylization intensity using temperature scaling. 
The style key features $\mathbf{K}_s$ are scaled by a user-controllable factor $\tau$ before attention, allowing fine-grained control over stylization strength.}
 \label{fig:ablation_study_kscaling}
\end{figure*}


\subsection{Controllable Stylization Strength}
\label{key_scaling}

In our Style Fusion Attention layers, the network merges content and style features with varying influence across different head regions. While a single style reference \(\mathrm{I}_{\text{style}}\) provides a target aesthetic, users may want to explore multiple levels of stylization for the same reference—ranging from subtle, content-preserving edits to strong, style-focused transformations.

To provide this flexibility, we introduce a temperature scaling mechanism in the Style Fusion Attention layers. The style key features \(\mathbf{K}_s\) are scaled by a user-controllable scalar factor \(\tau > 1\) prior to the Softmax operation, effectively adjusting the relative contribution of style features during attention computation. This scaling enables continuous control over the stylization strength, allowing the generation of multiple stylization intensities from a single style reference. 

The attention mechanism is computed as:

\[
\mathbf{Q} = \mathbf{Q}_l, \quad 
\mathbf{K} = [\mathbf{K}_l, \mathbf{\hat{K}}_c, \tau \mathbf{K}_s]^{\top}, \quad 
\mathbf{V} = [\mathbf{V}_l, \mathbf{V}_c, \mathbf{V}_s]^{\top}.
\]

This formulation allows the latent features to selectively attend to content and style information in a controllable manner, producing multiview-consistent stylizations that can be adjusted from subtle to strong intensity while preserving identity.
\begin{figure*}[t!]
    \centering
    \includegraphics[width=1.0\linewidth]{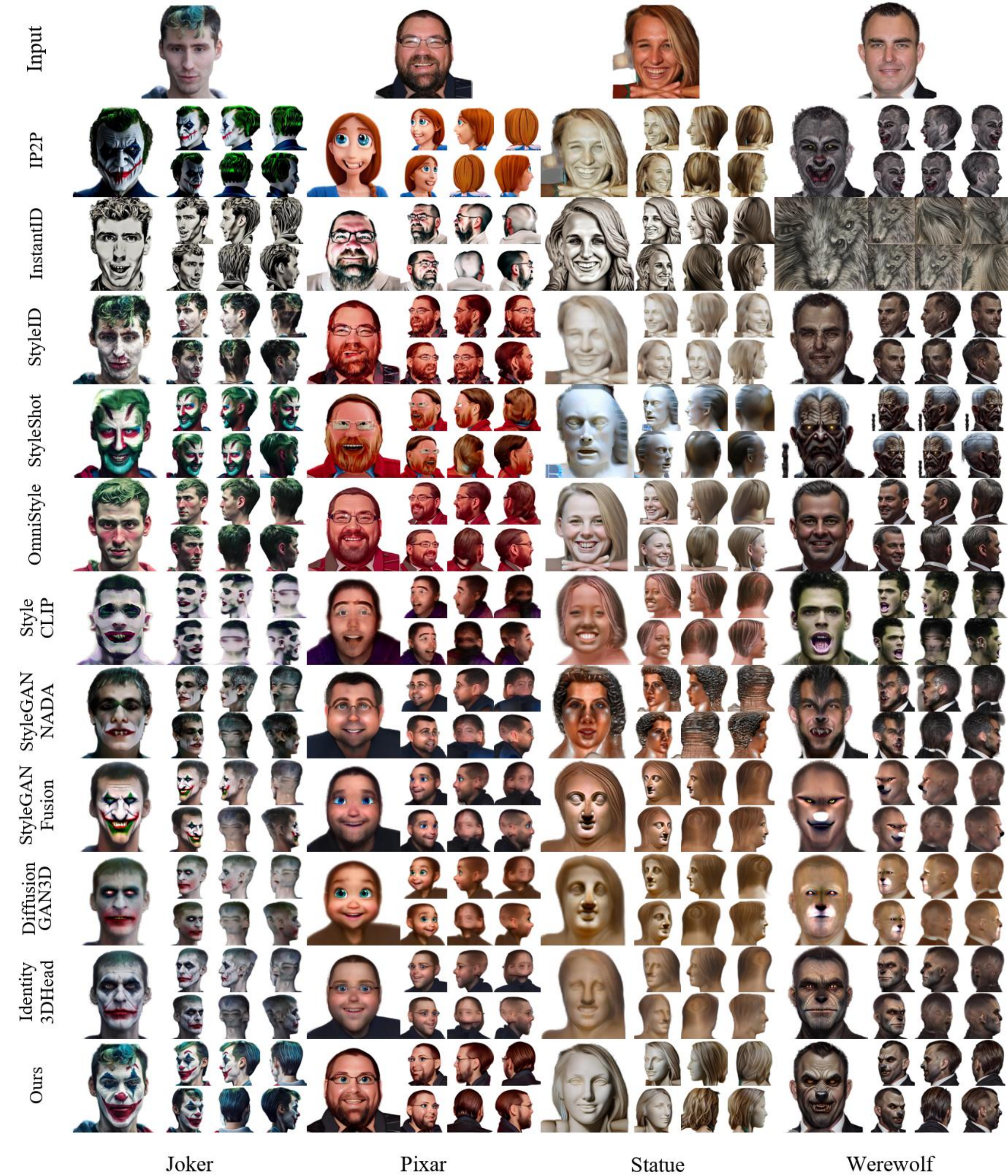}
      \vspace{-0.7cm}
    \caption{Qualitative stylization results of 3D head stylization methods, provided in 360-degree views.}
    \label{fig:results_visual}
\end{figure*}

\begin{table*}[t]
\centering
\resizebox{1.0\textwidth}{!}{
\begin{tabular}{lcccc|ccccc}
\toprule
 & \multicolumn{4}{c|}{\textbf{FFHQ }} & \multicolumn{5}{c}{\textbf{RenderMe360 }} \\
\cmidrule(lr){2-5} \cmidrule(lr){6-10}
\textbf{Method} 
& FID $\downarrow$ 
& CLIP $\uparrow$ 
& MEt3R $\downarrow$ 
& User Study $\uparrow$
& FID $\downarrow$ 
& CLIP $\uparrow$ 
& $\Delta\mathcal{D}$ $\downarrow$ 
& ID $\uparrow$ 
& MEt3R $\downarrow$ \\
\midrule
IP2P~\cite{brooks2023ip2p}                     & 87.00   & 0.7799 & 12.27 & 7.3\%  & 91.04 & \uline{0.7666} & 0.0965 & 0.24 & 18.51 \\
InstantID~\cite{wang2024instantid}            & 107.54  & 0.6966 & 14.40 & --     & 149.96 & 0.6440 & 0.1060 & 0.42   & 21.87 \\
StyleCLIP~\cite{patashnik2021styleclip}       & 126.22  & 0.7294 & 10.76 & --     & 124.23 & 0.6603 & 0.0365 & 0.27    & 16.60 \\
OmniStyle~\cite{wang2025omnistyle} & 121.37 & 0.7038 & 12.40 & - & 94.47 & 0.7101 & 0.0941 & 0.30 & 17.69 \\
StyleID~\cite{chung2024styleid}   & 109.30 & 0.7199 & 13.91 & - & \uline{90.58} & 0.7008 & 0.1034 & \uline{0.43} & 16.76 \\
StyleShot~\cite{gao2024styleshot} & \uline{86.23}  & 0.7390 & 13.92 & - & 109.22  & 0.7279 & 0.0973 & 0.15 & 20.66 \\
StyleGAN-Fusion~\cite{song2024DiffusionGuidedDomainAdaptation}  & 104.47  & 0.8031 & 11.65 & 2.7\%  & 96.95 & 0.6956 & \uline{0.0303}   & 0.32    & 16.87 \\
StyleGANNADA~\cite{gal2022stylegannada}       & 120.96  & 0.7607 & 14.20 & --     & 116.27 & 0.6842 & 0.0352   & 0.26       & 20.79 \\
DiffusionGAN3D~\cite{lei2024diffusiongan3d}      & 140.62  & 0.7802 & \textbf{9.68}  & --     & 113.04 & 0.6685 & 0.0307   & 0.36     & \textbf{15.20} \\
Identity3DHead~\cite{Bilecen2025IdentityPreserving3DHeadStylization} & 88.26   & \uline{0.8209} & 11.64 & \uline{14.0\%} & 111.01 & 0.7155 & 0.0325  & 0.31   & 16.44  \\
\textbf{Ours}                                   & \textbf{64.48} & \textbf{0.8237} & \uline{9.99 } & \textbf{76.0\%}  & \textbf{74.86} & \textbf{0.7939} & \textbf{0.0235}  & \textbf{0.45}    & \uline{15.55}  \\
\bottomrule
\end{tabular}}
\caption{Quantitative comparison on FFHQ and RenderMe360 datasets, including user study preference rates on FFHQ.}
\label{tab:avg_ffhq_renderme_combined}
\end{table*}

\section{Experiments}

\noindent \textbf{Metrics.}
We evaluate stylization performance using FID ~\cite{heusel2017gans} and CLIP embedding similarity. Identity preservation and multi-view consistency are assessed via the ID score \cite{deng2019arcface} and L2 depth difference ($\Delta\mathcal{D}$) between stylized multi-view images and their corresponding real multi-view images. Consistency is further evaluated using the MEt3R score~\cite{asim24met3r}. All methods are tested on the FFHQ~\cite{kerras2018stylegan} and RenderMe360~\cite{pan2024renderme} datasets.

For FID and CLIP evaluation, following \cite{lei2024diffusiongan3d}, we generate two distributions per style: one using the Stable Diffusion pipeline in the 2D image domain as style-specific reference images, and the other produced by the 3D head stylization methods. Example images from the reference distribution are provided in the Supplementary Material. We compute the FID between the two distributions and the CLIP similarity between corresponding image pairs.
While this protocol provides a practical and standardized benchmark for comparison, it is inherently limited by the absence of real paired multi-view stylization data.

The MEt3R score is calculated between consecutive camera views $(0,1)$, $(1,2)$, $\ldots$, $(N-1,N)$, as well as between the last and first view to close the loop. To improve readability, resulting scores are multiplied by 100.
The depth-based metric $\Delta\mathcal{D}$ is computed as the L2 difference between generated images and unedited multi-view ground-truth images from RenderMe360, using a depth estimation model~\cite{yang2024depthanything}.
We also conduct a user study to evaluate perceptual quality, the details of which are provided in the Supplementary Material.

\noindent  \textbf{Baselines.}
We compare our method with several 2D stylization and reference-based generation approaches, including InstructPix2Pix (IP2P)~\cite{brooks2023ip2p}, InstantID~\cite{wang2024instantid}, StyleID~\cite{chung2024styleid}, StyleShot~\cite{gao2024styleshot}, and OmniStyle~\cite{wang2025omnistyle}.
Following the pipeline established for 2D baselines, we first utilize the front-view content image \(\mathrm{I}_{\text{content}}\) to generate a stylized 2D image and then provide the resulting image to DiffPortrait360~\cite{gu2025diffportrait360} to generate stylized novel views.

We also compare our approach with several GAN-based face and head stylization methods, all adapted for the PanoHead generator. StyleCLIP~\cite{patashnik2021styleclip} trains a $\mathcal{W}^+$ mapper network using a CLIP-based loss. StyleGAN-NADA~\cite{gal2022stylegannada} optimizes selectively chosen generator layers with a CLIP objective instead of modifying $\mathcal{W}^+$. StyleGANFusion~\cite{song2024DiffusionGuidedDomainAdaptation} applies Score Distillation Sampling (SDS) for domain adaptation and introduces a directional regularizer through the frozen generator to achieve more stable training. DiffusionGAN3D~\cite{lei2024diffusiongan3d} enhances this regularization by incorporating a relative distance loss. Identity3DHead \cite{Bilecen2025IdentityPreserving3DHeadStylization} further improves identity preservation and stylization quality using negative log-likelihood distillation (LD).  
To test those methods that require GAN inversion, we optimize $\mathcal{W}^+$ to reconstruct the input image with the original PanoHead generator. The optimized $\mathcal{W}^+$ codes are then passed through the respective stylization generators to produce renderings from multiple viewpoints. Additional implementation details are provided in the Supplementary Material.

\noindent  \textbf{Results.}
We provide the  qualitative and quantitative results in Fig. \ref{fig:results_visual} and Table \ref{tab:avg_ffhq_renderme_combined}, respectively. 
As shown in the figure, our method produces the most realistic and view-consistent stylizations, and accurately reconstructs the back of the head. In contrast, GAN-based approaches consistently struggle to generate plausible back-view geometry and texture. Although we also make use of GAN-generated data during training, our data is generated by sampling rather than inferred from real images, and therefore does not exhibit the same back-view artifacts seen in GAN inference.
Moreover, applying a 2D stylization method followed by DiffPortrait360 yields noticeably inferior results compared to our method in both style fidelity and view consistency.

For quantitative evaluation, our method achieves the best FID and CLIP scores, indicating both high visual quality and strong style alignment. We also obtain the highest ID similarity score, demonstrating superior identity preservation compared to prior methods.
In terms of view consistency, we obtain comparable or better performance as measured by MEt3R. For MEt3R, we note that our objective is not to surpass GAN-based methods that explicitly model 3D structure using triplanes. Nevertheless, despite not employing any explicit 3D representation, our method achieves competitive multi-view consistency to these approaches.
Furthermore, when evaluating depth predictions derived from multi-view results on the RenderMe360 benchmark, our method outperforms GAN-based baselines. This suggests that our approach yields more accurate implicit 3D reconstruction from a single input image, even without relying on a 3D generator backbone.
Furthermore, based on a user study comparing our method against the three strongest baselines, participants significantly preferred our method, as shown in Table \ref{tab:avg_ffhq_renderme_combined}.

\newcommand{\vcenterimg}[2]{%
    \begin{tabular}[c]{@{}c@{}}\includegraphics[width=#1]{#2}\end{tabular}%
}

\newcommand{\ablationblock}[1]{%
\begin{tabular}[c]{@{}c@{}c@{}}
    \vcenterimg{0.12\linewidth}{#1/00.jpg} &
    \begin{tabular}[c]{@{}c@{}}
        \vcenterimg{0.06\linewidth}{#1/02.jpg} \\[-2pt]
        \vcenterimg{0.06\linewidth}{#1/14.jpg}
    \end{tabular}%
\end{tabular}%
}

\begin{figure}[t!]
    \centering
    \footnotesize
    \setlength{\tabcolsep}{1.5pt} 

    \begin{tabular}{cccccc}
        Input & Style & StyleAligned \cite{hertz2024stylealigned} & w/o AdaIN & w/o Attention & Ours \\
        \hline
        \noalign{\vskip 5pt}

        \vcenterimg{0.12\linewidth}{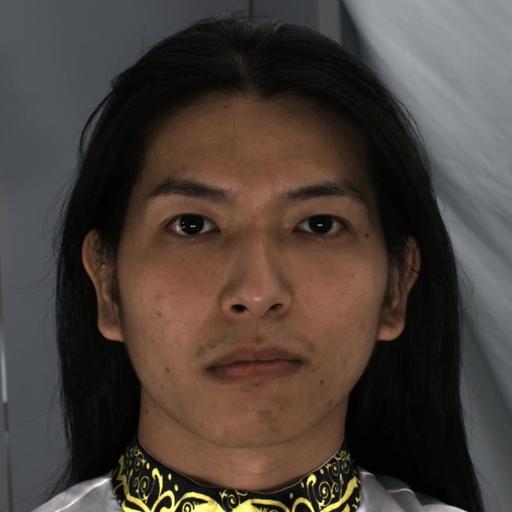} &
        \vcenterimg{0.12\linewidth}{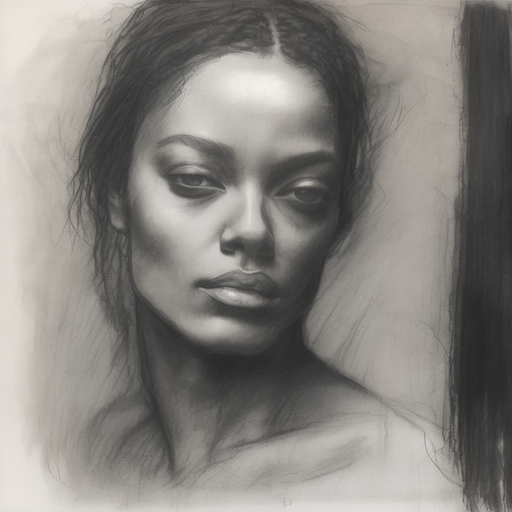} & 
        \ablationblock{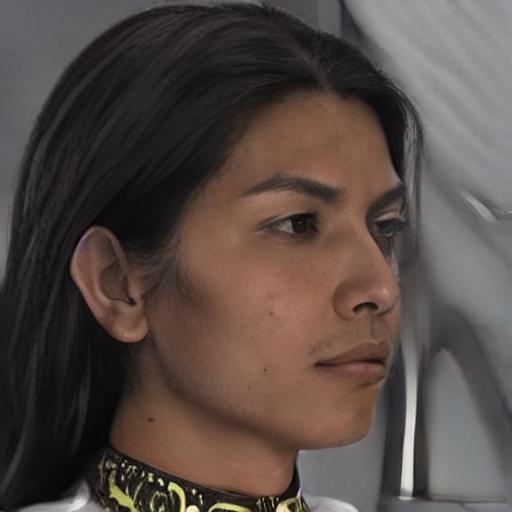} &
        \ablationblock{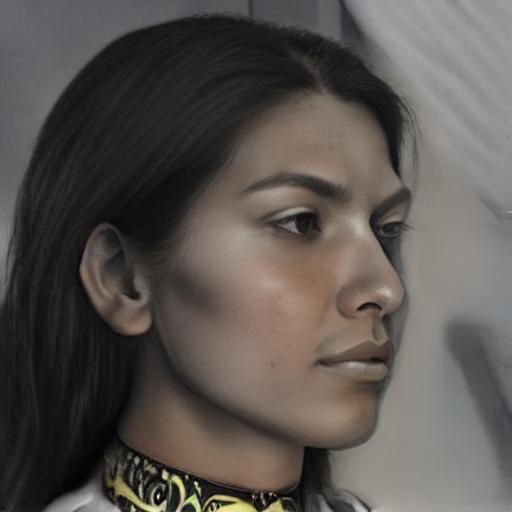} &
        \ablationblock{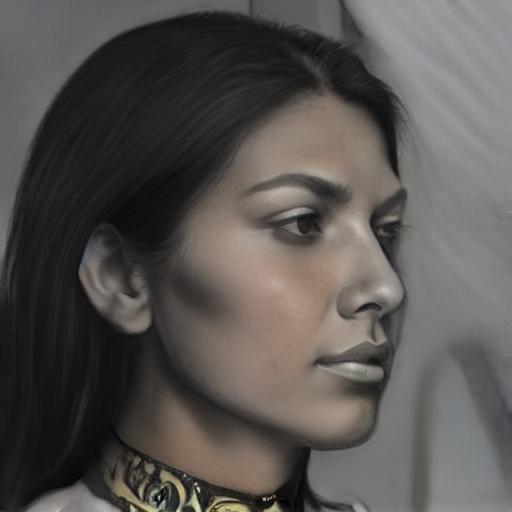} &
        \ablationblock{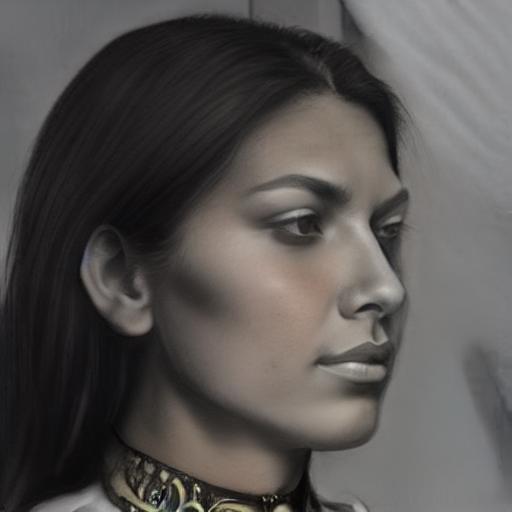} \\ 

        \vcenterimg{0.12\linewidth}{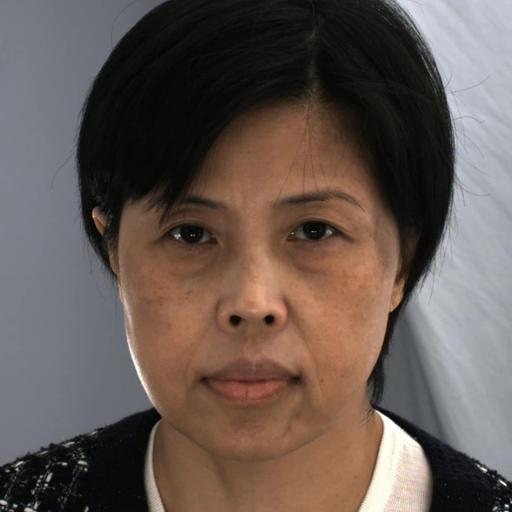} &
        \vcenterimg{0.12\linewidth}{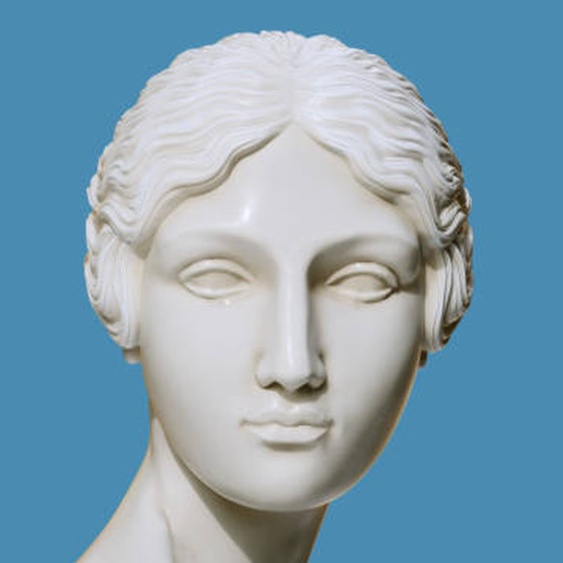} & 
        \ablationblock{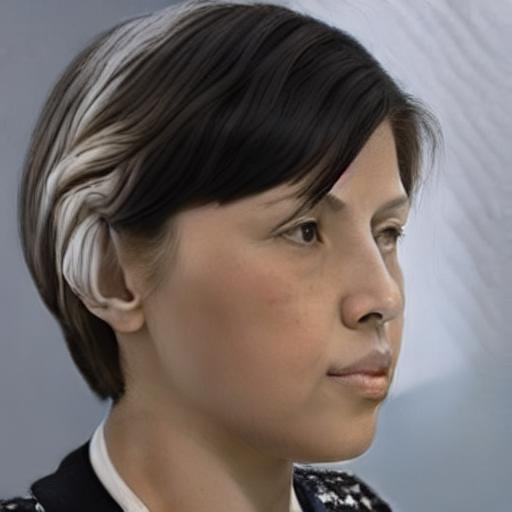} &
        \ablationblock{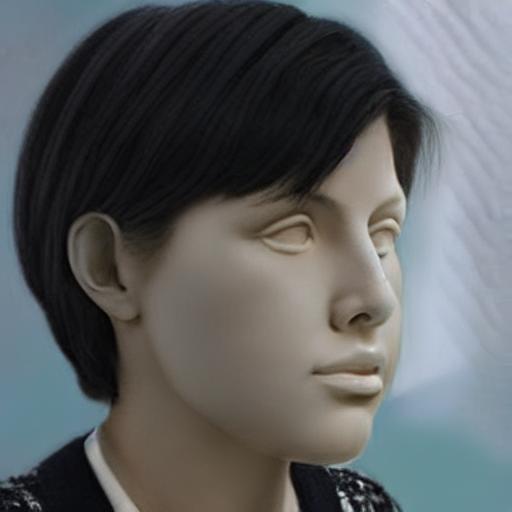} &
        \ablationblock{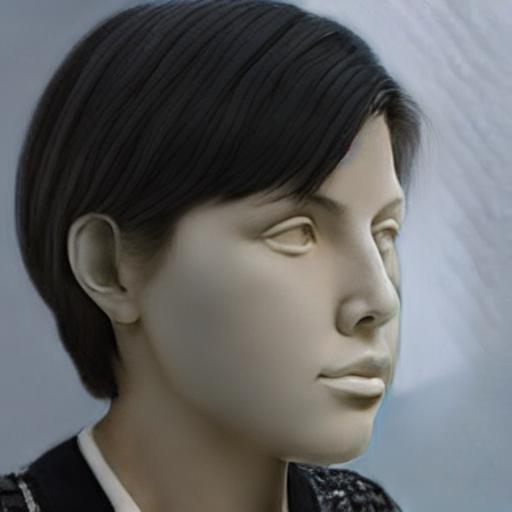} &
        \ablationblock{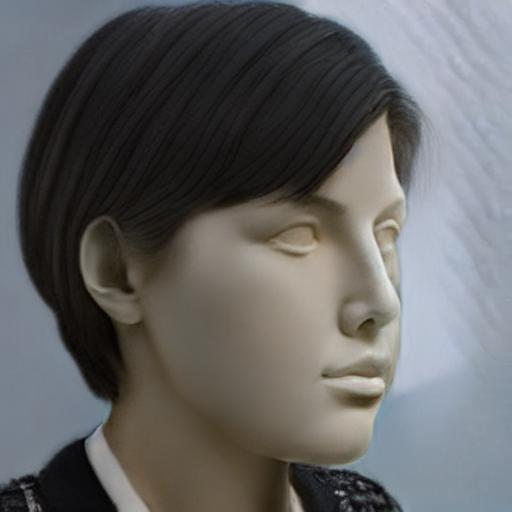} \\

        \vcenterimg{0.12\linewidth}{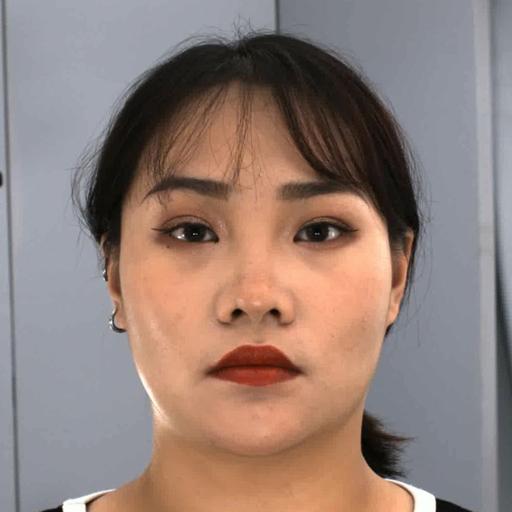} &
        \vcenterimg{0.12\linewidth}{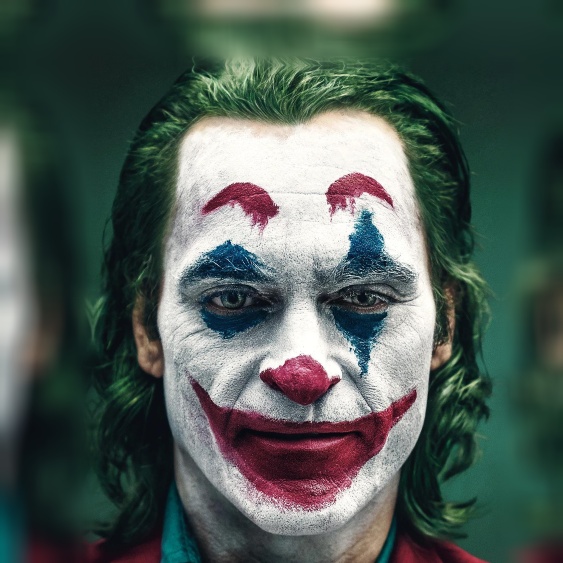} & 
        \ablationblock{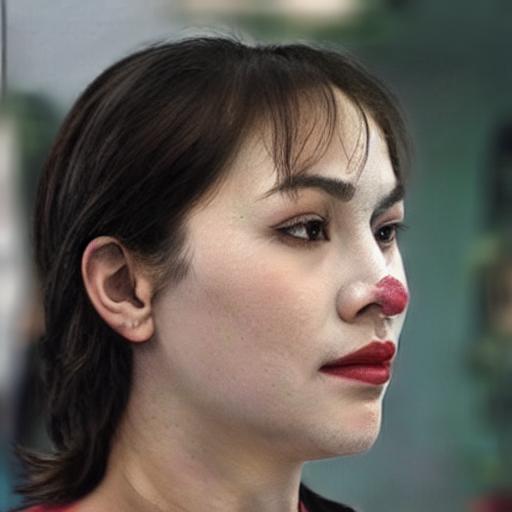} &
        \ablationblock{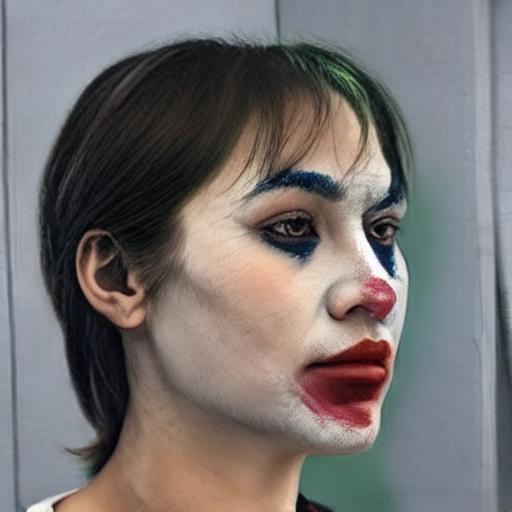} &
        \ablationblock{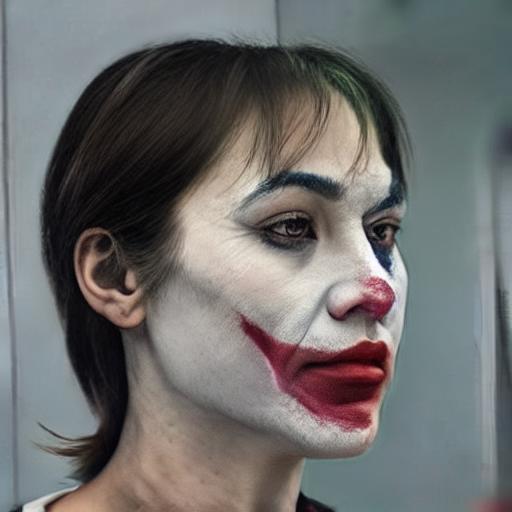} &
        \ablationblock{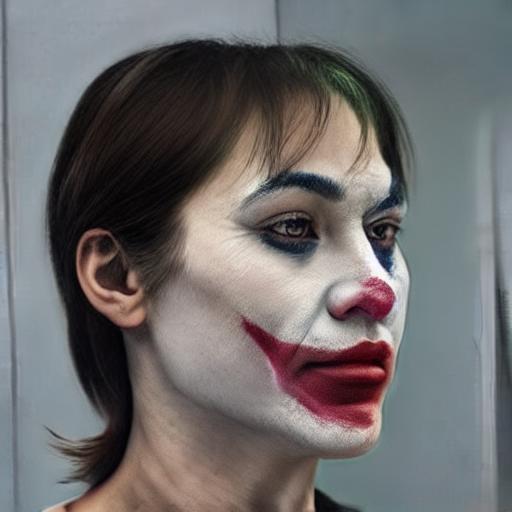} \\
    \end{tabular}

    \caption{Ablation study of the content key modulation components.
    Without AdaIN, the model fails to properly inject the target style and the outputs remain close to the input appearance. Without attention, partial stylization occurs but spatial consistency and structural coherence degrade. The full model produces stylizations that better reflect the reference style while preserving the input identity and multi-view consistency.}
    \label{fig:ablation_study_all}
    \vspace{-0.3cm}
\end{figure}

\noindent  \textbf{Ablation Study.}
In our analysis, we first evaluate StyleAligned~\cite{hertz2024stylealigned}. While effective for style-consistent image generation in 2D diffusion models, it performs poorly in our 3D-aware setting. Integrating StyleAligned into the pipeline fails to preserve multi-view consistency, often producing view-dependent artifacts and inconsistent style propagation across viewpoints.
To analyze the contribution of the main components of our framework, we then conduct an ablation study by removing the adaptive normalization (AdaIN) and the attention module separately. Fig.~\ref{fig:ablation_study_all} compares the stylization results under three settings: (i) without AdaIN in key modulation, (ii) without attention in key modulation, and (iii) the full model.
Removing AdaIN significantly weakens the style transfer capability. As shown in the second column of ablations, the generated outputs remain visually close to the input images and fail to reflect the stylistic characteristics of the reference image. This indicates that AdaIN plays a critical role in injecting global style statistics into the content representation.
When the attention-based modulation is removed, the model still transfers some stylistic cues but exhibits less coherent spatial stylization. This behavior is expected since the style and content images are not spatially aligned; without attention, the model cannot establish correspondences between their features, leading to less consistent region-wise modulation.
In contrast, the full model successfully combines both components to produce stylized outputs that closely follow the reference style while preserving the identity and structural details of the input. The results demonstrate that AdaIN is essential for effective style modulation, whereas the attention mechanism improves the spatial alignment and consistency of stylization.

We also experimented with applying the same modulation to the value features of the content branch. However, this modification produced results similar to those obtained with key modulation. Therefore, we omit it from the final design to avoid introducing additional computational overhead.



\begin{figure}
    \centering
    \includegraphics[width=0.8\linewidth]{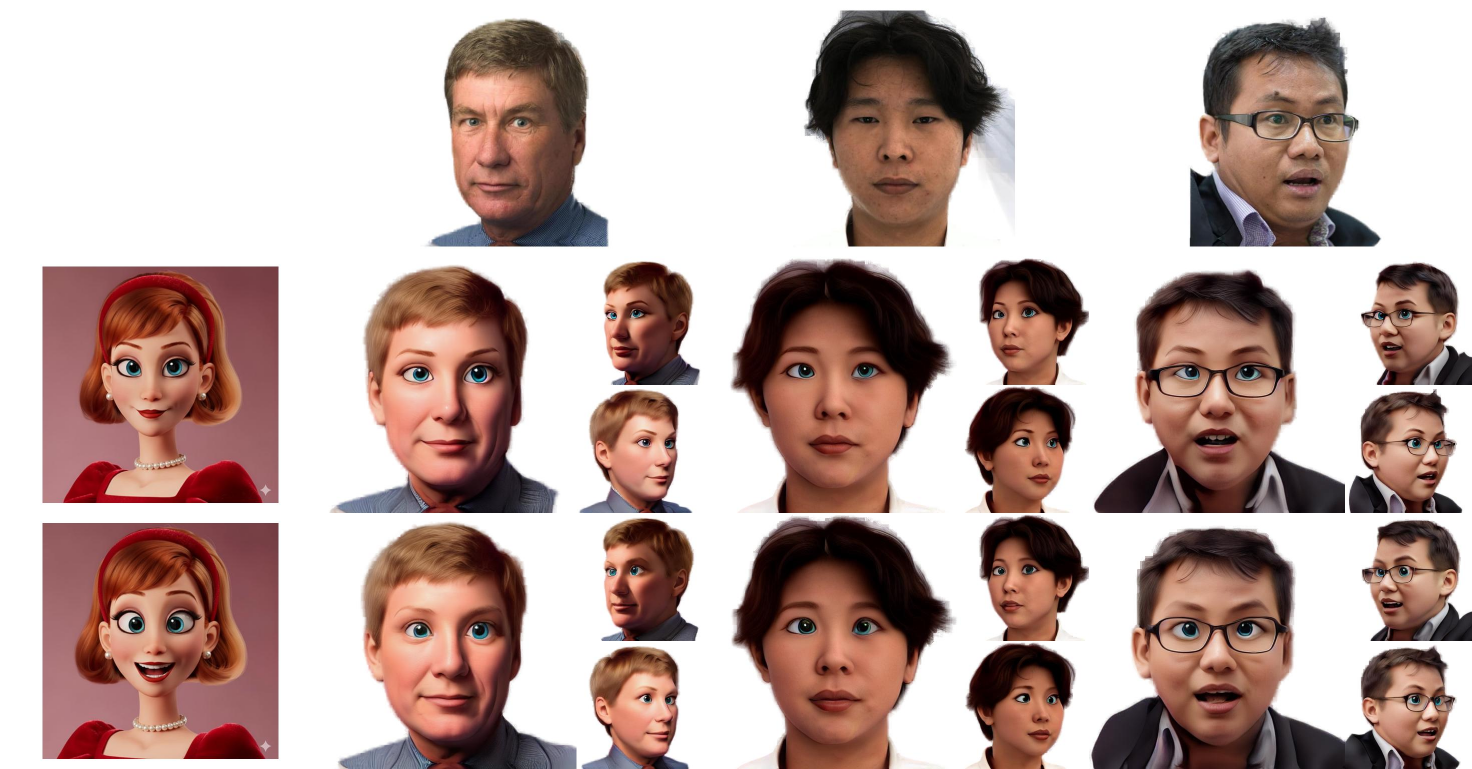}
\caption{Robustness to style expression variations. 
Altering the facial expression of the style image results in nearly identical stylized outputs, indicating that our method captures high-level stylistic attributes rather than transferring expression details.}
    \label{fig:roboust_expression}
\end{figure}

\noindent  \textbf{Robustness to Style Image Expression.}
We  analyze whether the facial expression in the style image influences the behavior of our method. 
Specifically, we construct a controlled experiment in which the expression of the style reference is deliberately modified while keeping all other factors unchanged. 
As shown in Fig. \ref{fig:roboust_expression},  when substantially altering the expression of the style image, we observe that the generated outputs remain nearly unchanged in overall appearance and stylistic interpretation. 
This indicates that our model does not directly transfer facial expressions from the style reference, but instead captures higher-level stylistic attributes, demonstrating robustness to expression variations.
Although certain geometric attributes (e.g., eye shape or size) may adapt to satisfy the target style (e.g., exaggerated proportions in a Pixar-like stylization), these modifications are consistent and identity-preserving. Core identity-related characteristics remain stable across expression variations, indicating that the model disentangles expression from stylistic representation.

\begin{figure}[t!]
  \centering
      \includegraphics[width=\linewidth]{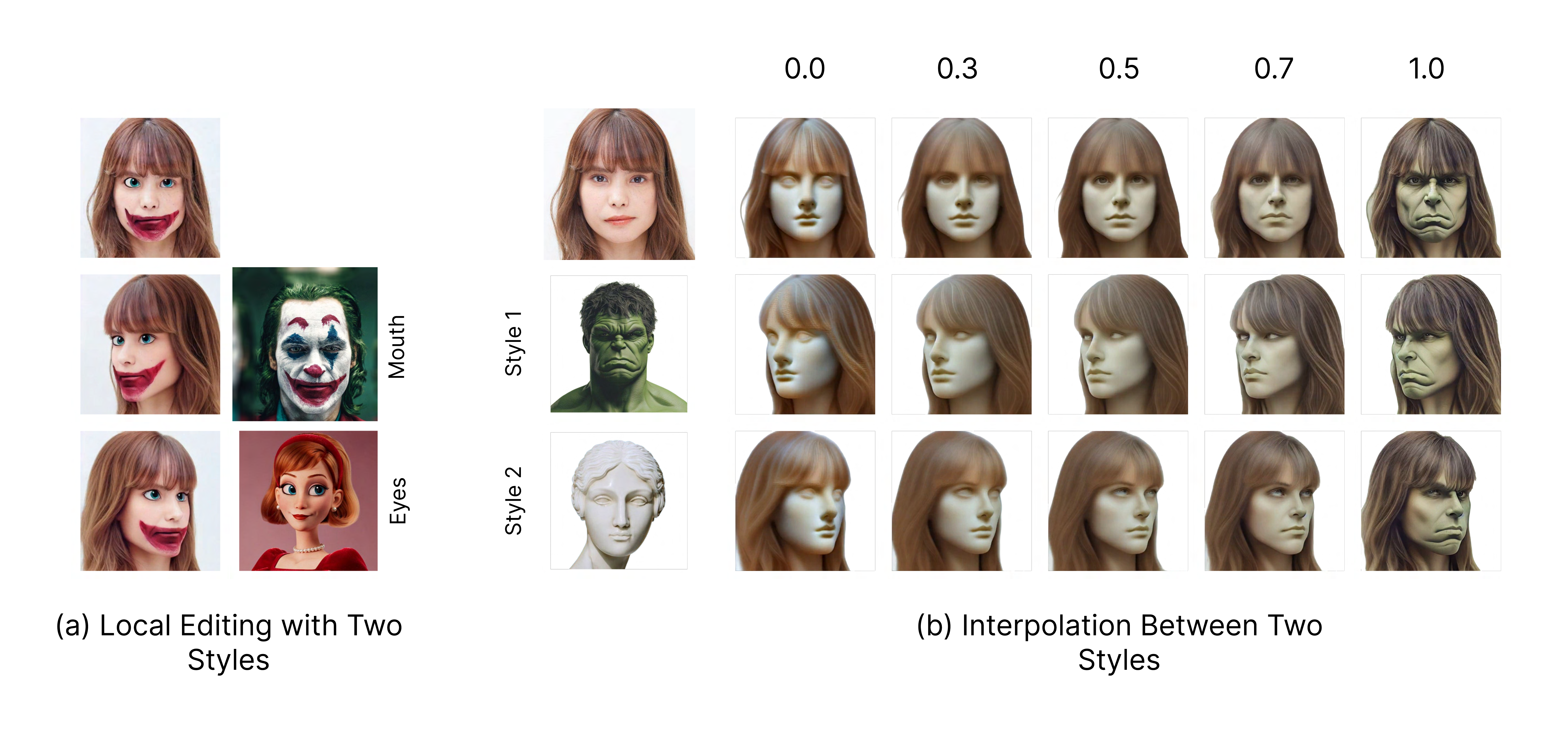}
  \vspace{-1cm}
  \caption{(a) Local Multi-Editing. (b) Visualization of linear interpolation between two styles. The columns show the generated results as the interpolation ratio $\alpha$ transitions from Style 1 ($\alpha=0$) to Style 2 ($\alpha=1$).}
  \label{fig:interpolation}
\end{figure}

\noindent  \textbf{Local Stylization Results.}
In teaser, \cref{fig:teaser}, presents few results of our local stylization method, demonstrating that stylization can be applied to selected regions of the head while preserving the appearance of the other areas.
We also show that our method can achieve multiple localized edits simultaneously. For example, as illustrated in \cref{fig:interpolation}(a), we can stylize the mouth region using a “Joker” style while applying a “Pixar” style to the eyes. This flexibility highlights the capability of our style mask approach to combine multiple styles within a single image.

\noindent  \textbf{Style Interpolation Results.}
To blend two distinct styles, we apply our feature extraction process to two style reference images, yielding style appearance features $\mathbf{F}_{s_1}$ and $\mathbf{F}_{s_2}$. The final style key $\mathbf{K}_s$ and value $\mathbf{V}_s$ are computed by projecting each feature set and then linearly interpolating the results with a blending factor $\alpha \in [0, 1]$. This process is defined as:
\[
\begin{aligned}
    \mathbf{K}_s &= (1 - \alpha) (W_K \mathbf{F}_{s_1}) + \alpha (W_K \mathbf{F}_{s_2}), \\
    \mathbf{V}_s &= (1 - \alpha) (W_V \mathbf{F}_{s_1}) + \alpha (W_V \mathbf{F}_{s_2}).
\end{aligned}
\]
This formulation provides fine-grained control over the stylistic output, allowing for a smooth and continuous transition between the two styles by simply adjusting the scalar value of $\alpha$, as visualized in Figure~\ref{fig:interpolation}(b).

\newcommand{\imgwidth}{0.14\linewidth}   

\newcommand{\cellimg}[1]{%
  \begin{subfigure}[t]{\imgwidth}
    \centering
    \includegraphics[width=\linewidth]{#1}
  \end{subfigure}%
}
\section{Conclusion}

We presented \textbf{StyleFusion360}, a diffusion-based framework for  3D head stylization from a single input and style reference. By introducing the Style Appearance Module for disentangled style transfer, the Style Fusion Attention for controllable stylization strength, and multi-view diffusion fine-tuning for cross-view coherence, our approach achieves high-quality stylization without requiring retraining for new styles. Extensive evaluations demonstrate that StyleFusion360 outperforms existing baselines in terms of style fidelity, identity preservation, and geometric consistency.
As future work, recent methods \cite{liao2025soap, zhou2025zero} could be incorporated into pose-conditioned attention layers to maintain structural and stylistic consistency across expression and pose variations.

\bibliographystyle{splncs04}
\bibliography{main}

\title{Supplementary Material for StyleFusion360: View-Consistent Head Stylization via Adaptive Style Modulation}

\titlerunning{StyleFusion360 Supplementary Material}

\author{Furkan Güzelant, Arda Göktoğan, Tarık Kaya, and Ayşegül Dündar}

\authorrunning{F.~G\"uzelant et al.}

\institute{Bilkent University, Ankara, Turkey \\
\email{\{furkan.guzelant, arda.goktogan, tarik.kaya\}@bilkent.edu.tr, adundar@cs.bilkent.edu.tr}}

\maketitle

\section{Dataset Pair Examples}

\begin{figure*}[t!]
    \centering
    \includegraphics[width=1.0\linewidth]{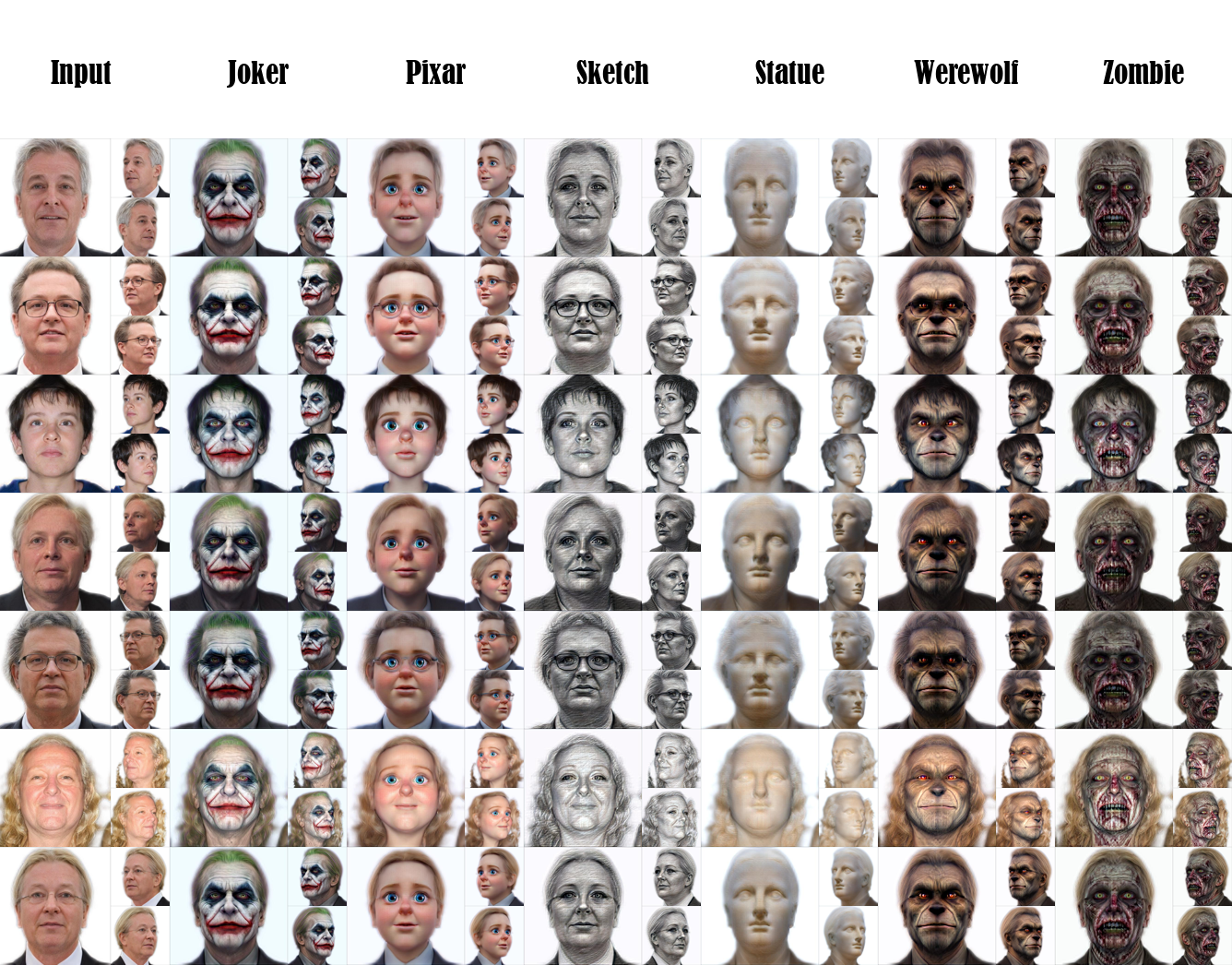}
      \vspace{-0.7cm}
    \caption{Dataset pair examples.}
    \label{fig:dataset_styles}
\end{figure*}

\cref{fig:gan_dataset_pairs} presents an example triplet from the \textit{Joker} style domain, consisting of a content image generated by a base 3D-GAN generator~\cite{an2023panohead}, a cross-identity style exemplar, and the corresponding ground-truth stylized output produced by a domain-adapted 3D-GAN model~\cite{Bilecen2025IdentityPreserving3DHeadStylization}. 
Note that the style exemplar itself is also generated using the same domain-adapted model. 
Using a style image from a different identity encourages a clearer separation between identity and style, enabling more effective disentanglement of content and stylistic attributes during training. 
For our dataset, we generate 150 samples for each of the six style domains, yielding a total of 900 training pairs. A representative subset of these samples is shown in \cref{fig:dataset_styles}.

\newcommand{\vcenterimggan}[2]{%
    \begin{tabular}[c]{@{}c@{}}\includegraphics[width=#1]{#2}\end{tabular}%
}

\newcommand{\gandataset}[1]{%
\begin{tabular}[c]{@{}c@{}c@{}}
    \vcenterimggan{0.20\linewidth}{supplementary/gan_dataset/#1/view_00.png} &
    \begin{tabular}[c]{@{}c@{}}
        \includegraphics[width=0.10\linewidth]{supplementary/gan_dataset/#1/view_03.png} \\[-1.5pt]
        \includegraphics[width=0.10\linewidth]{supplementary/gan_dataset/#1/view_29.png}
    \end{tabular}%
\end{tabular}%
}

\begin{figure}[hbt!]
    \centering
    \footnotesize
    \setlength{\tabcolsep}{1.5pt} 

    \begin{tabular}{ccc}
        \textbf{Content Image} & \textbf{Style Image} & \textbf{Ground Truth} \\
        \hline
        \noalign{\vskip 5pt}

        \gandataset{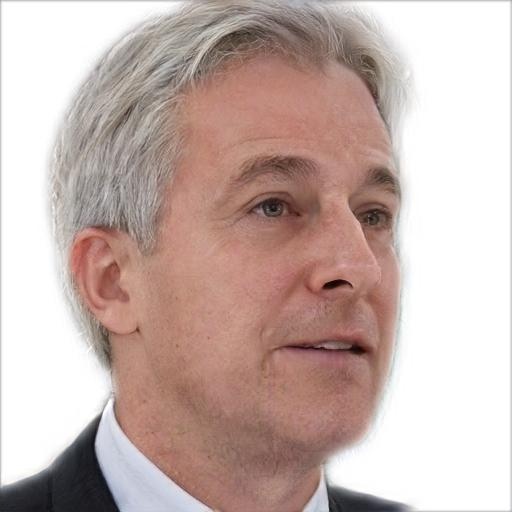} &
        \gandataset{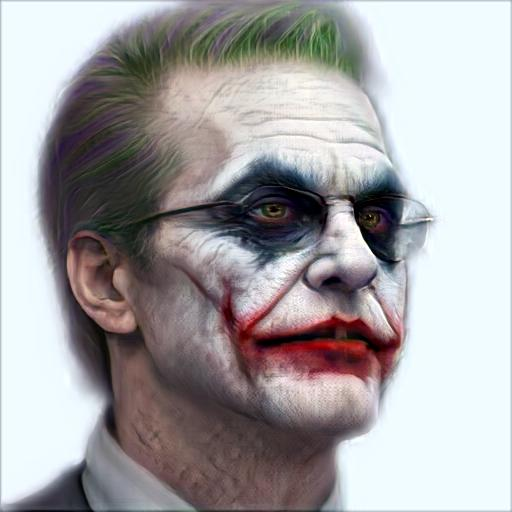} &
        \gandataset{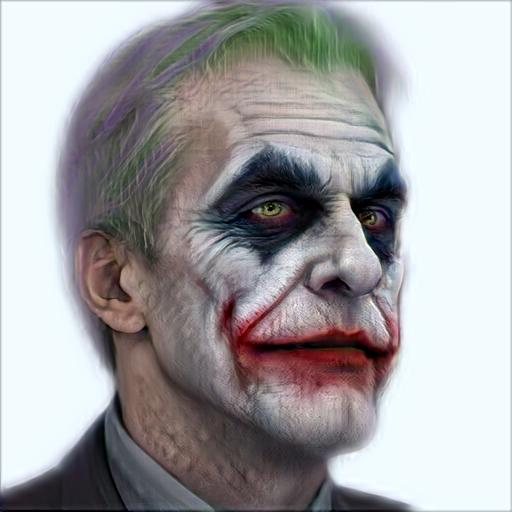} \\
    \end{tabular}
    
    \caption{Dataset pairs used in our training.}
    \label{fig:gan_dataset_pairs}
    \vspace{-0.3cm}
\end{figure}

\section{Implementation Details}

\subsection{Training Setup} 

We fine-tune the Style Appearance and View Consistency modules of our model on a synthetic 3D-GAN-generated stylized dataset for 800 iterations. The dataset includes six style categories: Joker, Pixar, Sketch, Statue, Werewolf, and Zombie. These domains are selected to cover a diverse range of stylistic transformations, including color-dominant, geometry-altering, texture-heavy, and non-photorealistic styles. We use the AdamW optimizer with a learning rate of $10^{-5}$ and apply Classifier-Free Guidance (CFG) with a weight of 3.0. The batch size is 16, corresponding to the number of generated views per generation. All experiments are conducted on a single NVIDIA A100 GPU. We use $\tau = 1.05$ as the default key scaling value unless otherwise specified. During training, style references are sampled from identities different from the content image to encourage disentanglement between identity and style. For local stylization, we obtain region masks using an off-the-shelf segmentation model, SAM2~\cite{ravi2024sam2}. The masks are converted into binary maps for the target regions, resized to match the spatial resolution of the attention features, and applied to the style key features in the Style Fusion Attention layers. For multi-style local editing, different region masks are associated with different style reference images and applied within the same diffusion pass.

\subsection{Baselines}

For diffusion-based stylization methods (IP2P~\cite{brooks2023ip2p}, InstantID~\cite{wang2024instantid}, StyleID~\cite{chung2024styleid}, StyleShot~\cite{gao2024styleshot}, and OmniStyle~\cite{wang2025omnistyle}), we first stylize the input content image using a text-driven style prompt. The resulting stylized image is then passed through DiffPortrait360~\cite{gu2025diffportrait360} to synthesize multi-view stylized outputs.

We train the latent mapper of StyleCLIP \cite{patashnik2021styleclip} using the PanoHead generator. For StyleGAN-NADA \cite{gal2022stylegannada} and StyleGANFusion \cite{song2024DiffusionGuidedDomainAdaptation}, we rely on the official implementation of StyleGANFusion and adapt the generator backbone to PanoHead. In the case of StyleGANFusion, we use their EG3D configuration for PanoHead, and for StyleGAN-NADA \cite{gal2022stylegannada} we incorporate adaptive layer selection as described in the original method. DiffusionGAN3D \cite{lei2024diffusiongan3d} is implemented directly from the paper, as no official code is available. We preserve each baseline's original hyperparameters, such as denoiser settings, noise schedule, learning rate, and optimizer, unless training instability requires adjustments.

\subsection{Quantitative Scores}
\begin{figure}[t!]
    \centering
    \footnotesize
    \setlength{\tabcolsep}{1pt}

    \scalebox{1.0}{
        \begin{tabular}{ccc}
            \textbf{Ours} & \textbf{Input} & \textbf{Ground Truth} \\
            \hline
            \noalign{\vskip 3pt}
            \includegraphics[width=0.30\linewidth]{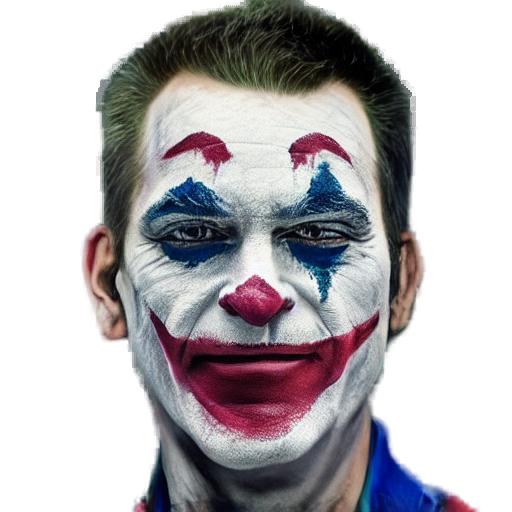} &
            \includegraphics[width=0.30\linewidth]{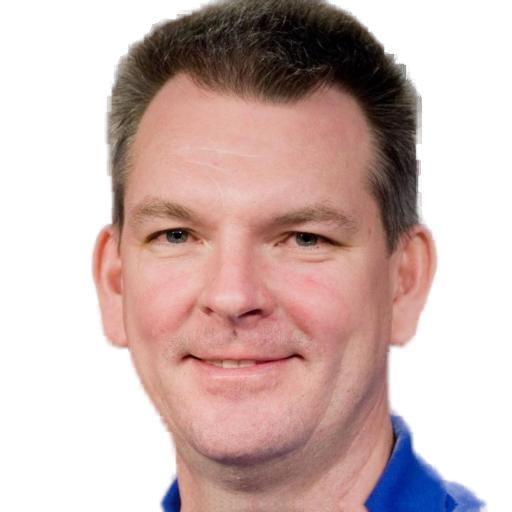} &
            \includegraphics[width=0.30\linewidth]{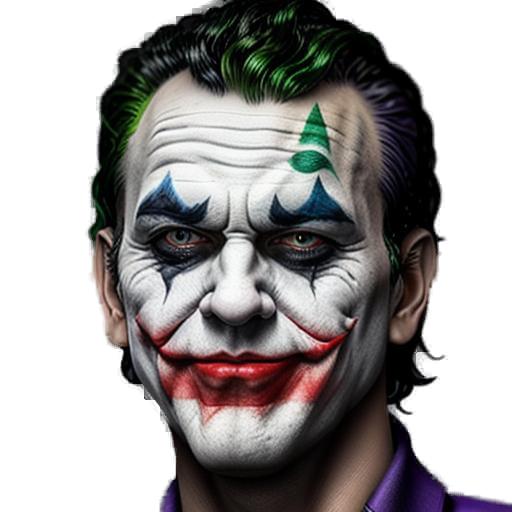} \\

            \includegraphics[width=0.30\linewidth]{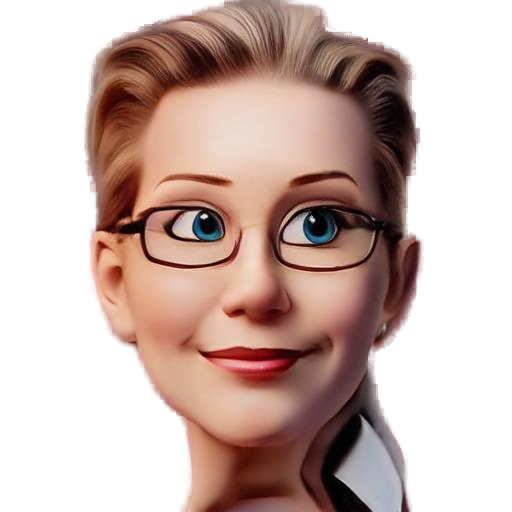} &
            \includegraphics[width=0.30\linewidth]{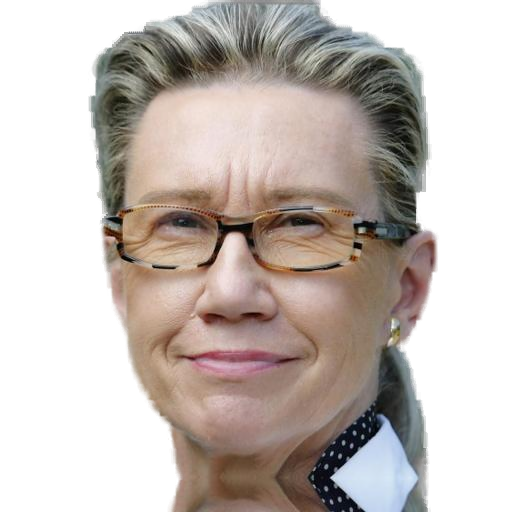} &
            \includegraphics[width=0.30\linewidth]{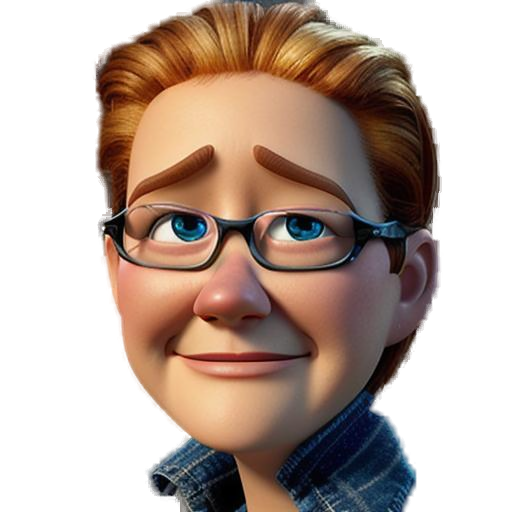} \\

        \end{tabular}
    }
    
    \caption{From left to right: a stylized image from the second distribution (outputs of our model), the corresponding unedited input image from the third distribution (used for evaluating identity preservation), and the edited result from the first distribution obtained by using the Stable Diffusion pipeline (used as the ground truth for FID and CLIP evaluations).
    \label{fig:quantitative_eval_example}
}
    \vspace{-0.3cm}
\end{figure}

To evaluate our approach, we generate three reference image distributions using the Stable Diffusion pipeline. 
The first distribution is created by adding noise at timestep $t=25$ to the input images and denoising them for 50 steps with the corresponding style prompt using each baseline's diffusion checkpoints. 
The resulting edited images serve as ground-truth targets for computing FID and CLIP similarity. 
The second distribution consists of the stylized outputs produced by the 3D head stylization methods, which we compare against the first distribution for FID and against matched pairs for CLIP scores. 
The third distribution contains the original, unedited images and is used to evaluate identity preservation (ID) and $\Delta D$. 
For ID preservation, we compute ArcFace-based~\cite{deng2019arcface} identity similarity between the stylized outputs and their corresponding input images. 
For depth consistency, $\Delta D$ is measured using depth maps estimated by the DepthAnything model~\cite{yang2024depthanything}. 
These metrics are computed by comparing images from the second and third distributions. 
\cref{fig:quantitative_eval_example} shows representative examples from all three distributions.

\subsection{Runtime and Memory Usage} 

\begin{table}[t]
\centering
\caption{Detailed breakdown of runtime components and peak GPU memory comparison on a single NVIDIA A100 GPU.}
\label{tab:detailed_runtime_memory}
\resizebox{0.95\linewidth}{!}{%
\begin{tabular}{llcc}
\toprule
\textbf{Method} & \textbf{Component / Phase Breakdown} & \textbf{Runtime} & \textbf{Peak GPU Memory} \\ \midrule

\multirow{2}{*}{Identity3DHead} & Per-Style Optimization (Training) & 3 hours &  \\
 & Inference (Generation) & 2 sec & 2.11 GiB \\ \midrule

\multirow{2}{*}{DiffusionGAN3D} & Per-Style Optimization (Training) & 50 mins &  \\
 & Inference (Generation) & 2 sec & 1.49 GiB \\ \midrule

\multirow{5}{*}{2D Stylization + DiffPortrait360} & 2D Stylization Step (Average) & 20 sec &  \\
 & 3D-Aware Noise Initialization & 15 sec &  \\
 & Back-Head Generation & 30 sec &  \\
 & 360° Multiview Synthesis & 90 sec & 40.5 GiB \\
 & \textbf{Total Inference} & \textbf{2 mins 35 sec} &  \\ \midrule

\multirow{6}{*}{\textbf{Ours}} & Per-Style Optimization (One-time, 6 styles) & 18 hours &  \\
 & Dataset Generation (One-time) & 10 mins &  \\
 & Model Fine-Tuning (One-time Training) & 2 hours &  \\
 & Inference: 3D-Aware Noise Initialization & 15 sec &  \\
 & Inference: Back-Head Generation (Content+Style) & 60 sec &  \\
 & Inference: 360° Multiview Synthesis & 120 sec & \textbf{45.8 GiB} \\
 & \textbf{Total Inference} & \textbf{3 mins} &  \\ 
\bottomrule
\end{tabular}%
}
\end{table}

All reported runtimes and memory measurements are obtained on a single NVIDIA A100 GPU. We report the inference runtime and peak GPU memory usage in \cref{tab:detailed_runtime_memory}. For methods that require per-style optimization or training, the runtime is reported per target style. For diffusion-based 2D baselines, the reported runtime includes both frontal-view stylization and subsequent multi-view synthesis using DiffPortrait360.

StyleFusion360 has a one-time training cost, after which it can stylize new content images with arbitrary style references without per-style retraining. This makes it substantially more flexible than GAN-based approaches that require optimization or adaptation for each target style. Although our method uses more GPU memory due to the diffusion-based multi-view generation backbone, its inference time remains comparable to the 2D stylization plus DiffPortrait360 pipeline while providing substantially better multi-view consistency.

\section{Additional Ablation Study}

\subsection{Key and Value Modulation}

\begin{figure}[t!]
\centering
\includegraphics[width=0.98\linewidth]{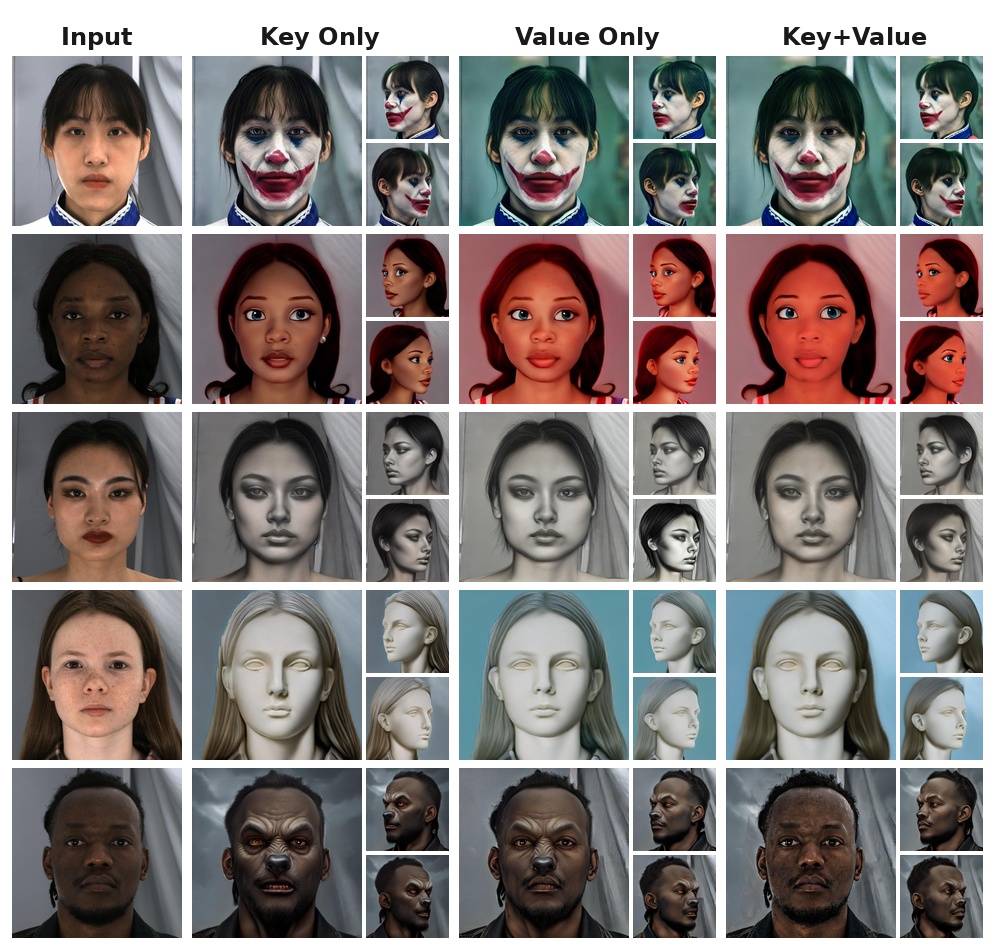}
\vspace{-2mm}
\caption{\textbf{Qualitative ablation of Style Fusion Attention.} Comparison between key-only (our method), value-only, and key+value modulation across diverse styles, including Joker, Pixar, Sketch, Statue, and Werewolf. Key-only modulation provides a better balance between style transfer and 3D structural consistency. In contrast, value-only modulation often introduces overly strong appearance changes, such as color oversaturation in Pixar, while key+value modulation may amplify stylization at the cost of identity and geometric stability.}
\label{fig:ablation_visual}
\vspace{-3mm}
\end{figure}

We further analyze where style-conditioned modulation should be applied within Style Fusion Attention. In addition to our key-only modulation design, we evaluate value-only modulation and joint key-value modulation. We provide both qualitative and quantitative comparisons in \cref{fig:ablation_visual} and \cref{tab:key_value_ablation}, respectively.

\begin{table}[t]
\centering
\caption{Quantitative ablation of key-only, value-only, and key+value modulation. Key-only modulation achieves the best FID and $\Delta D$, indicating stronger visual quality and better 3D consistency.}
\label{tab:key_value_ablation}
\begin{tabular}{lcccc}
\toprule
\textbf{Method} & \textbf{FID} $\downarrow$ & \textbf{CLIP} $\uparrow$ & \textbf{$\Delta D$} $\downarrow$ & \textbf{ID} $\uparrow$ \\
\midrule
Key Only    & \textbf{71.685} & 0.7939 & \textbf{0.0234} & 0.4245 \\
Value Only  & 84.602 & 0.7894 & 0.0253 & \textbf{0.4881} \\
Key + Value & 88.220 & \textbf{0.7968} & 0.0259 & 0.4625 \\
\bottomrule
\end{tabular}
\end{table}

As shown in \cref{fig:ablation_visual}, the key-only variant produces more stable stylization across viewpoints while better preserving the underlying 3D head structure. This behavior is especially visible in geometry-sensitive styles such as Statue and Werewolf, where value-only and key+value modulation tend to distort facial structure or produce less coherent side views. For appearance-heavy styles such as Pixar, value-only modulation transfers color and texture strongly, but often leads to oversaturation and weaker identity preservation. Key+value modulation can further increase the apparent style strength, but it also makes the generation less stable and may overemphasize style-specific appearance changes.

The quantitative results in \cref{tab:key_value_ablation} support these observations. Key-only modulation achieves the best FID and $\Delta D$, showing that it provides the strongest balance between visual quality and 3D consistency. Value-only modulation obtains the highest ID score, but its worse FID and $\Delta D$ indicate weaker visual realism and geometric stability. Key+value modulation slightly improves CLIP similarity, suggesting stronger style alignment, but it performs worse in FID and $\Delta D$. This indicates that stronger style injection comes at the cost of visual quality and multi-view structural consistency. Therefore, we adopt key-only modulation as our final design.

\section{User Study}
We conducted a user study with \textbf{50 participants}, each of whom completed 
\textbf{18 comparison questions}. Options were randomly permuted to ensure fairness. In each question, the participant was shown:
\begin{itemize}
    \item the original content image,
    \item a style exemplar image,
    \item stylized outputs produced by different methods.
\end{itemize}

Participants were asked to select the image that provides the \textbf{best balance of 
identity preservation and stylization}, \ie, the image that both retains recognizable 
identity and reflects the target style. We did not ask separate 
questions for identity and stylization; participants provided a single holistic judgment for each comparison.

The stylization results are provided in gif format. 

\begin{table}[t]
\centering
\caption{User study preference on the FFHQ dataset, aggregated over 900 responses 
(50 participants $\times$ 18 questions). Higher values indicate stronger user preference.}
\label{tab:userstudy}
\begin{tabular}{l c}
\toprule
\textbf{Method} & \textbf{User Study (\%)} \\
\midrule
IP2P~\cite{brooks2023ip2p} & 7.3\% \\
StyleGAN-Fusion~\cite{song2024DiffusionGuidedDomainAdaptation} & 2.7\% \\
Bilecen \etal~\cite{Bilecen2025IdentityPreserving3DHeadStylization} & 14.0\% \\
\textbf{Ours} & \textbf{76.0\%} \\
\bottomrule
\end{tabular}
\end{table}

As shown in \cref{tab:userstudy}, participants preferred our method in 
\textbf{76\%} of all comparisons, indicating that our approach achieves the most favorable perceptual trade-off between identity retention and stylistic fidelity.

\section{Limitations}

Our framework is designed to transfer style in a semantically meaningful manner by aligning facial regions between the input identity and the reference style image. In particular, the model is encouraged to adopt stylistic attributes from corresponding regions of the style image (e.g., eye appearance, eyebrow shape, or lip texture). As a result, the method implicitly assumes style images that contain human facial structures. While this design choice enables more coherent and region-aware stylization for portrait styles, applying styles derived from non-human subjects may lead to less meaningful correspondences. Extending the approach to support more diverse style domains remains an interesting direction for future work.

\section{Evaluation Protocol and Limitations}
Since there is no real paired multi-view stylization dataset, quantitative evaluation of 3D head stylization remains challenging. Following recent related works such as DiffusionGAN3D~\cite{lei2024diffusiongan3d} and Identity3DHead~\cite{Bilecen2025IdentityPreserving3DHeadStylization}, we use Stable Diffusion-generated stylized images as style-specific reference distributions for FID and CLIP evaluation. Specifically, the reference distribution is obtained by applying a Stable Diffusion editing pipeline to the input images using the corresponding style prompt. The outputs of each 3D head stylization method are then compared against this reference distribution. This protocol provides a practical proxy for measuring visual quality and style alignment in the absence of real paired multi-view stylization ground truth. However, it is not a perfect substitute for human-created multi-view stylized targets. Therefore, we complement the quantitative metrics with a user study and extensive qualitative comparisons. We consider the construction of real paired multi-view stylization benchmarks an important direction for future work.

\section{Additional Generalization Results}

\begin{figure*}[t!]
    \centering
    \includegraphics[width=1.0\linewidth]{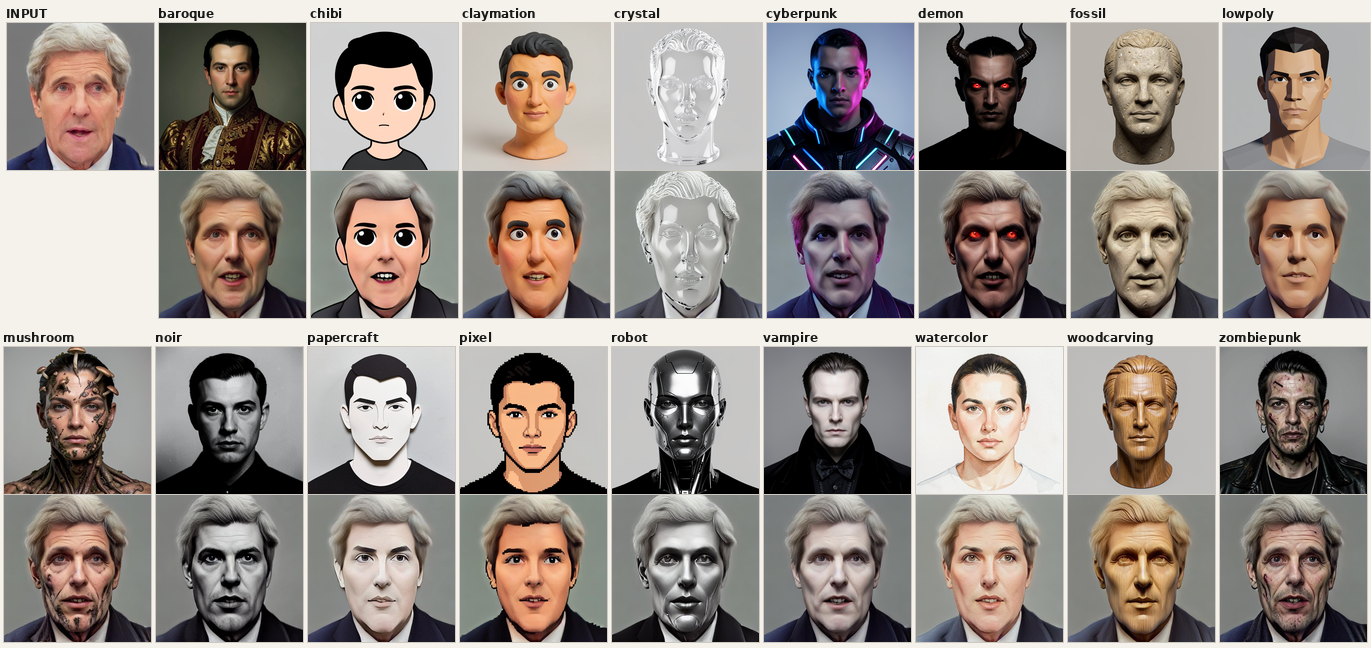}
    \vspace{-0.5cm}
    \caption{Out-of-domain stylization results on unseen style domains. Each column shows a reference style image and the corresponding output generated by our method from the same input identity. StyleFusion360 generalizes well to a diverse set of unseen styles while preserving identity.}
    \label{fig:outofdomain}
\end{figure*}

To further evaluate the generalization ability of StyleFusion360, we present additional results on style domains that are not included in the training set. As shown in \cref{fig:outofdomain}, our method can successfully transfer a wide range of unseen styles, including baroque, chibi, claymation, crystal, cyberpunk, demon, fossil, lowpoly, mushroom, noir, papercraft, pixel, robot, vampire, watercolor, woodcarving, and zombiepunk. Despite the large stylistic diversity, the generated outputs remain visually coherent and preserve the identity of the input subject.

We also provide a quantitative out-of-domain evaluation in \cref{tab:out_of_domain}. Our method achieves the best performance across all reported metrics, outperforming both StyleGAN-Fusion and Identity3DHead. These results further demonstrate that StyleFusion360 is not limited to a closed set of training styles and can generalize effectively to novel artistic domains without per-style retraining.

\begin{table}[h]
\centering
\caption{Out-of-domain stylization results on unseen style domains. Each column shows a reference style image and the corresponding output generated by our method from the same input identity. StyleFusion360 generalizes well to a diverse set of unseen styles while preserving identity.}
\label{tab:out_of_domain}
\begin{tabular}{lccc}
\hline
\textbf{Method} & \textbf{FID} $\downarrow$ & \textbf{$\Delta D$} $\downarrow$ & \textbf{ID} $\uparrow$  \\ \hline
StyleGAN-Fusion & 102.9677     & 0.14136             & 0.1388      \\
Identity3DHead  & 79.9965      & 0.02785             & 0.3168      \\
\textbf{Ours}            & \textbf{64.9917}      & \textbf{0.02354}             & \textbf{0.5335}      \\ \hline
\end{tabular}
\end{table}

\section{Additional Results}


\newcommand{\localedit}[2]{%
\begin{tabular}[c]{@{}c@{}c@{}}
    \vcenterimg{0.16\linewidth}{figures/local_edit/#1/#2/00.jpg} &
    \begin{tabular}[c]{@{}c@{}}
        \vcenterimg{0.08\linewidth}{figures/local_edit/#1/#2/03.jpg} \\[-2pt]
        \vcenterimg{0.08\linewidth}{figures/local_edit/#1/#2/29.jpg}
    \end{tabular}%
\end{tabular}%
}

\begin{figure}[t!]
    \centering
    \footnotesize
    \setlength{\tabcolsep}{0.5pt}

    \begin{subfigure}{\linewidth}
        \centering
        \begin{tabular}{ccccc}
            Style Image & Input & Output & Input & Output \\
            \hline
            \noalign{\vskip 5pt}

            \vcenterimg{0.16\linewidth}{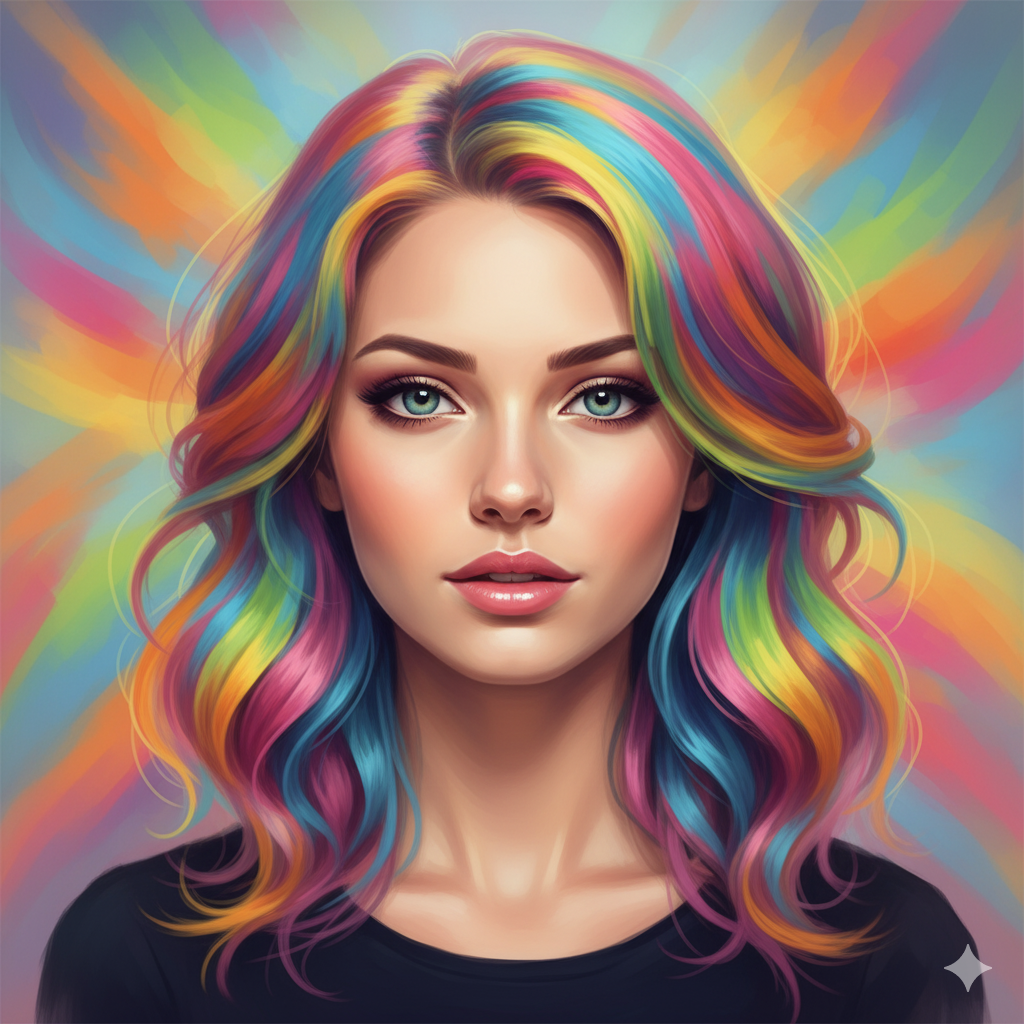} &
            \vcenterimg{0.16\linewidth}{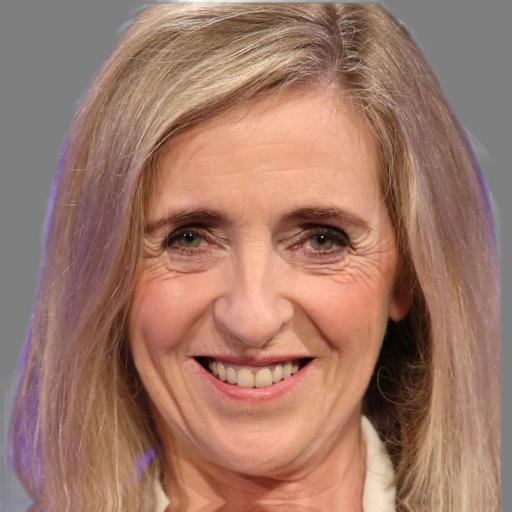} &
            \localedit{colorful_hair}{00251} &
            \vcenterimg{0.16\linewidth}{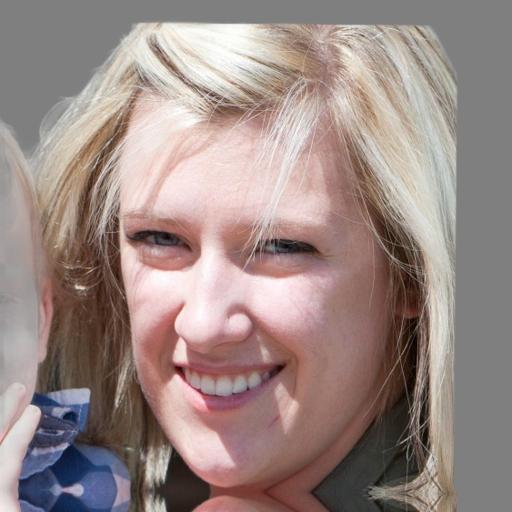} &
            \localedit{colorful_hair}{00138} \\ \noalign{\vskip 4pt}

            \vcenterimg{0.16\linewidth}{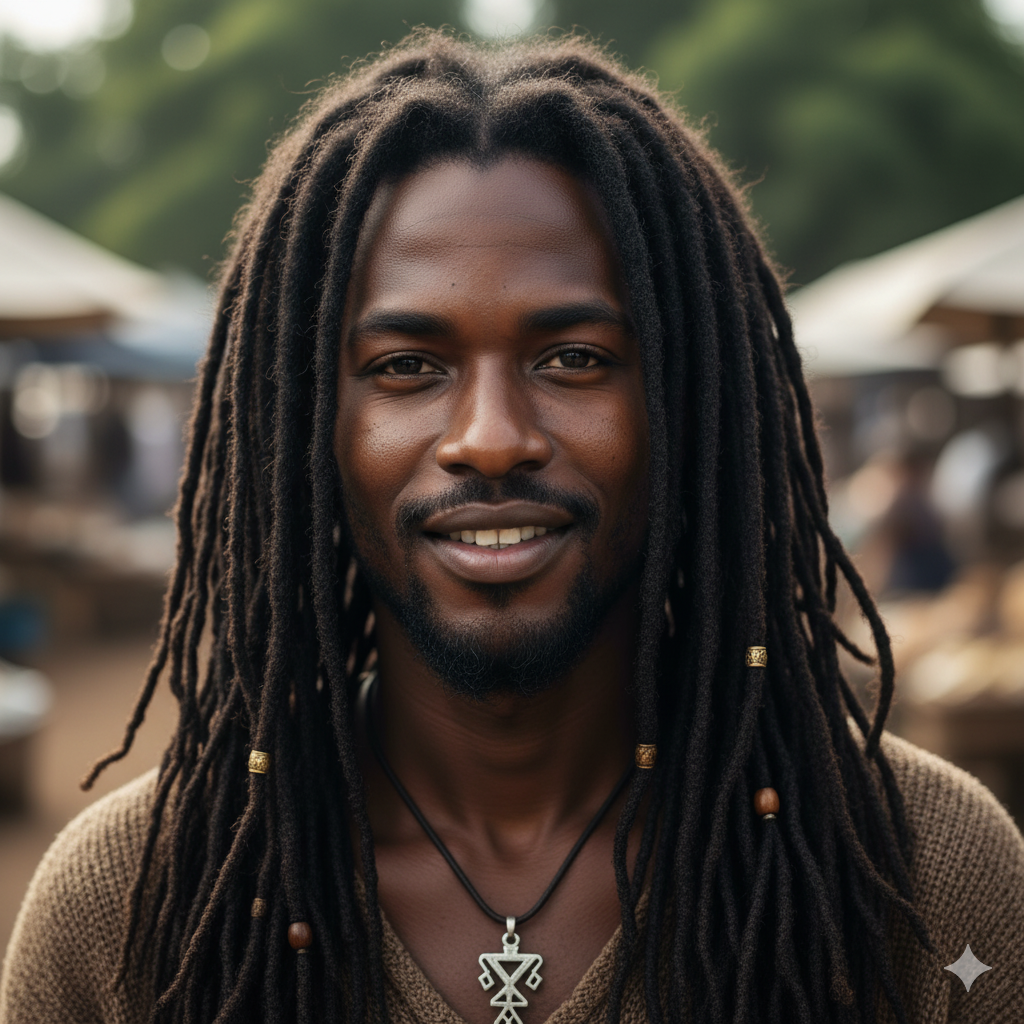} &
            \vcenterimg{0.16\linewidth}{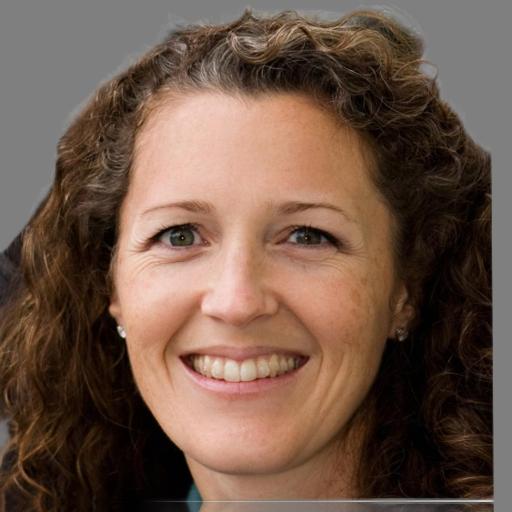} &
            \localedit{dreadlocks}{00207} &
            \vcenterimg{0.16\linewidth}{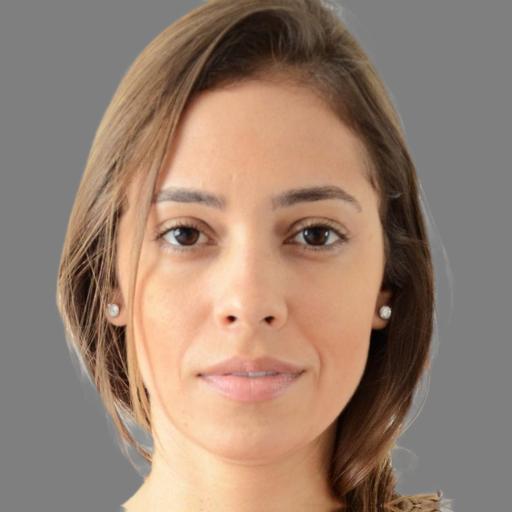} &
            \localedit{dreadlocks}{00200} \\ \noalign{\vskip 4pt}

            \vcenterimg{0.16\linewidth}{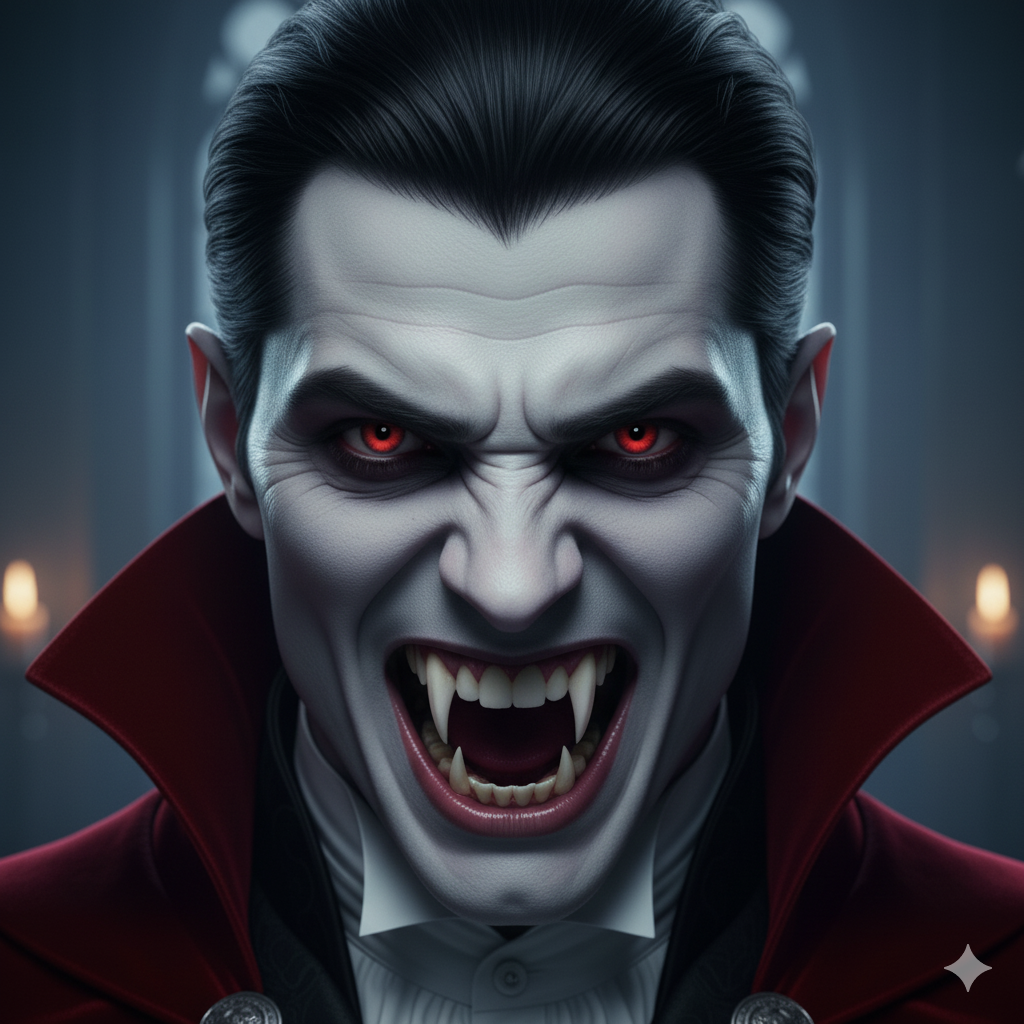} &
            \vcenterimg{0.16\linewidth}{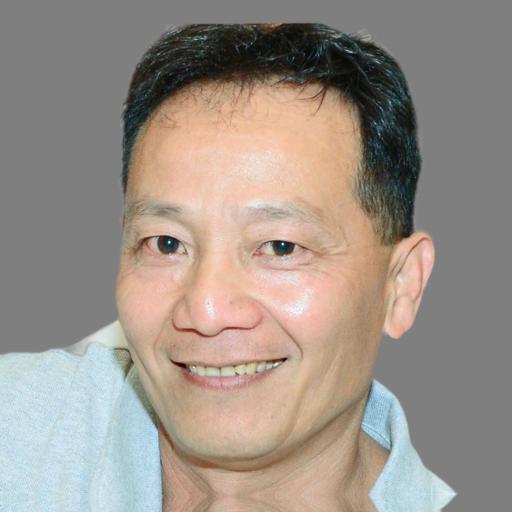} &
            \localedit{vampire}{00176} &
            \vcenterimg{0.16\linewidth}{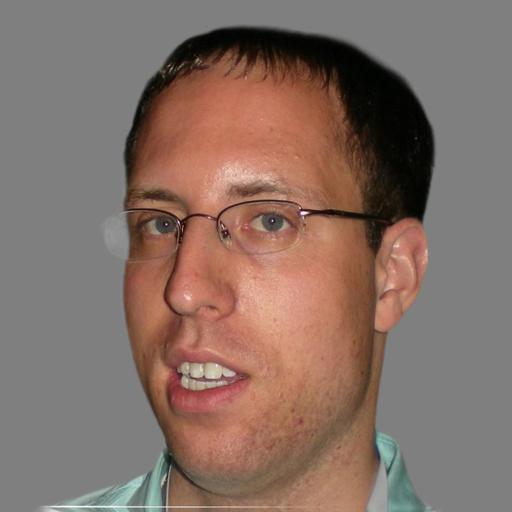} &
            \localedit{vampire}{00235} \\ \noalign{\vskip 4pt}

            \vcenterimg{0.16\linewidth}{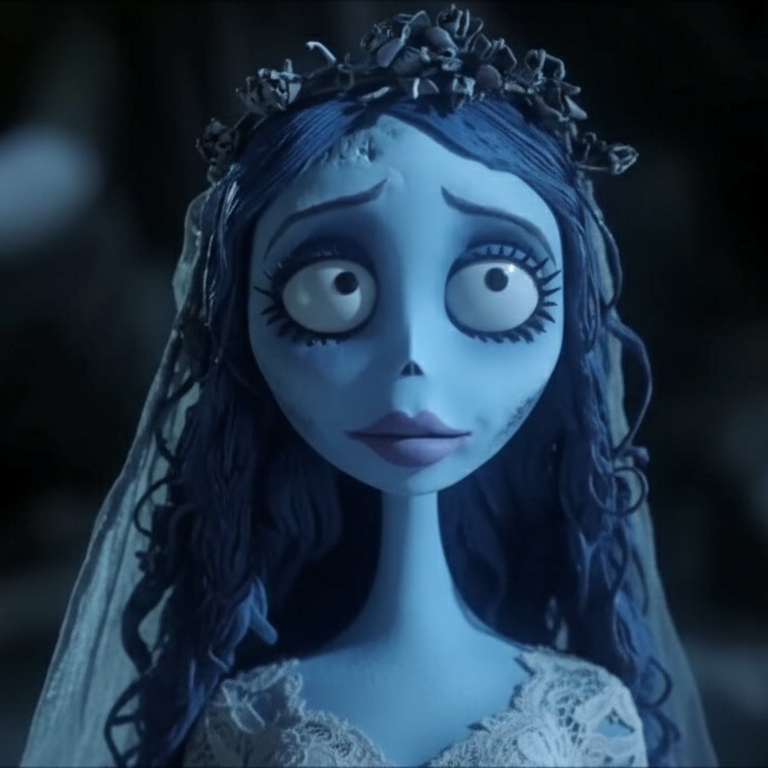} &
            \vcenterimg{0.16\linewidth}{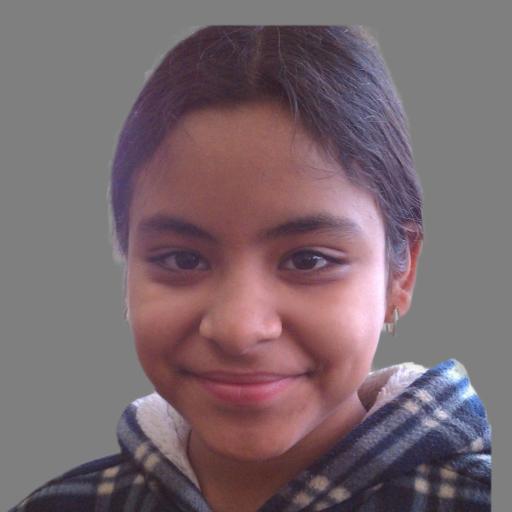} &
            \localedit{corpse_bride}{00116} &
            \vcenterimg{0.16\linewidth}{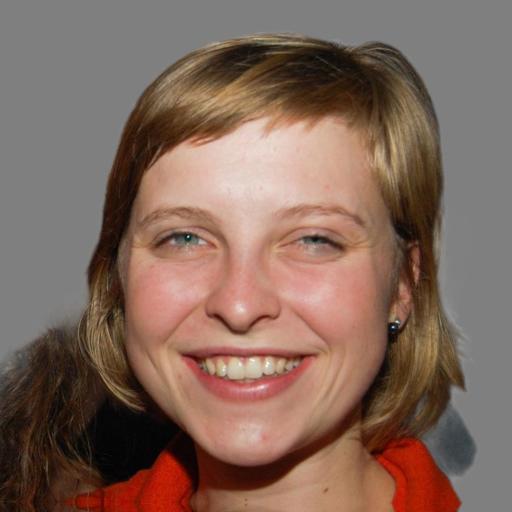} &
            \localedit{corpse_bride}{00167} \\
        \end{tabular}
        \caption{Local stylization results across different styles.}
        \label{fig:local_style_a}
    \end{subfigure}

    \vspace{0.4cm}

    \begin{subfigure}{\linewidth}
        \centering
        \begin{tabular}{cccc}
            Input & Style 1 & Style 2 & Output \\
            \noalign{\vskip 3pt}
            \vcenterimg{0.20\linewidth}{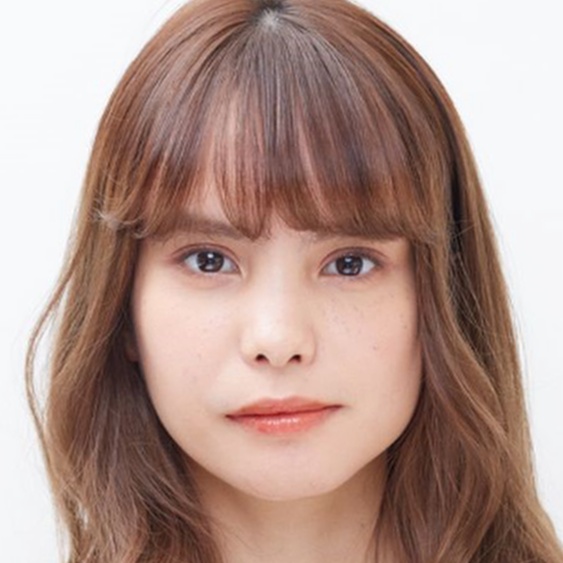} &
            \vcenterimg{0.20\linewidth}{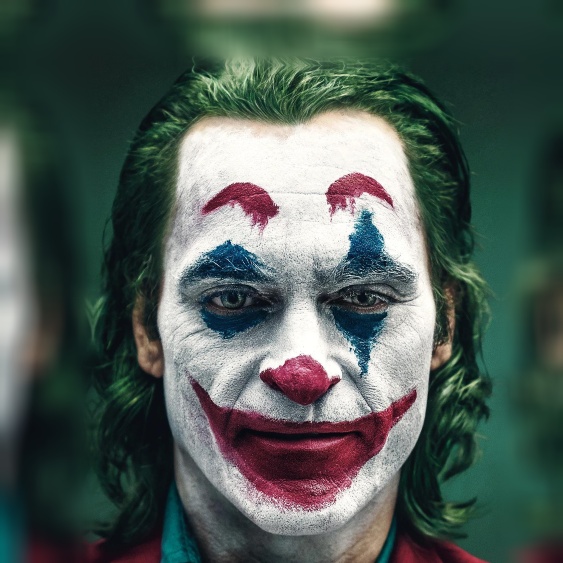} &
            \vcenterimg{0.20\linewidth}{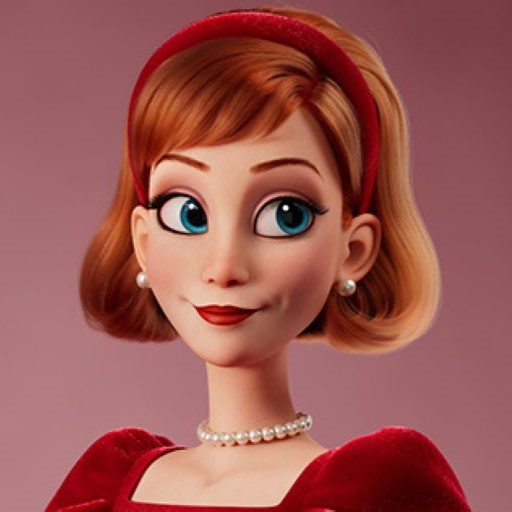} &
            \begin{tabular}[c]{@{}c@{}c@{}}
                \vcenterimg{0.20\linewidth}{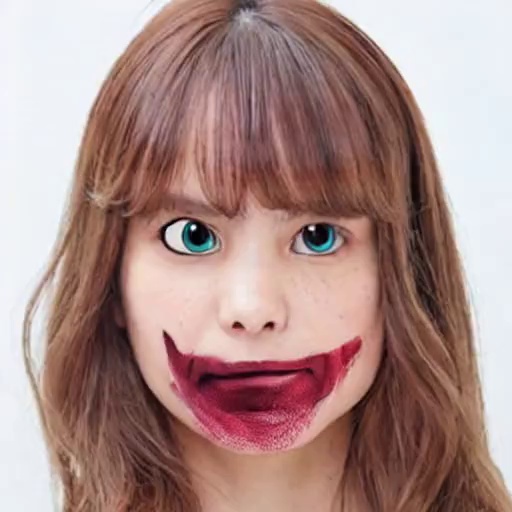} &
                \begin{tabular}[c]{@{}c@{}}
                    \vcenterimg{0.1\linewidth}{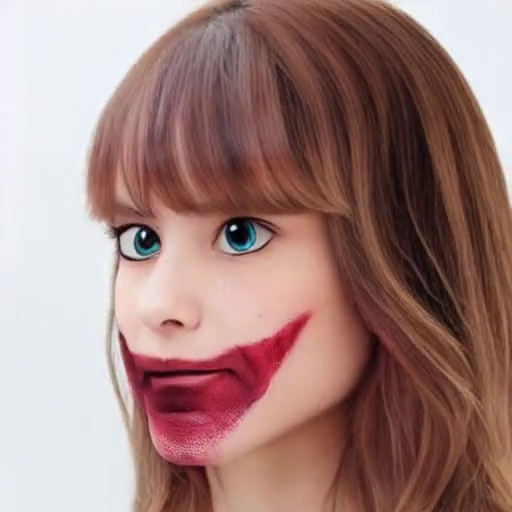} \\[-3pt]
                    \vcenterimg{0.1\linewidth}{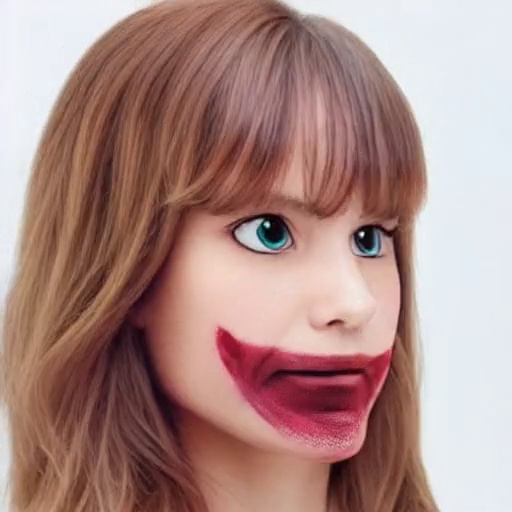}
                \end{tabular}%
            \end{tabular} \\
        \end{tabular}
        \caption{Example of combining two different style references.}
        \label{fig:local_style_b}
    \end{subfigure}
    \caption{(a) Local stylization results. (b) Multi-style local editing.}
    \label{fig:local_style_combined}
    
\end{figure}

In addition to global head stylization, our method supports fine-grained local edits and multi-style fusion. As demonstrated in \cref{fig:local_style_a}, the model can successfully isolate and transfer specific stylistic attributes from a reference image, such as colorful hair, dreadlocks, vampire fangs, or distinct eye makeup, while preserving the unedited regions of the original identity and maintaining strict multi-view consistency. Furthermore, \cref{fig:local_style_b} highlights the flexibility of our approach in combining multiple style references. By conditioning the generation on both a Joker and a Pixar style image, the model synthesizes a coherent output that seamlessly blends attributes from both domains, such as the exaggerated eye structure of the Pixar style and the characteristic mouth paint of the Joker.

\begin{figure*}[t!]
    \centering
    \includegraphics[width=1.0\linewidth]{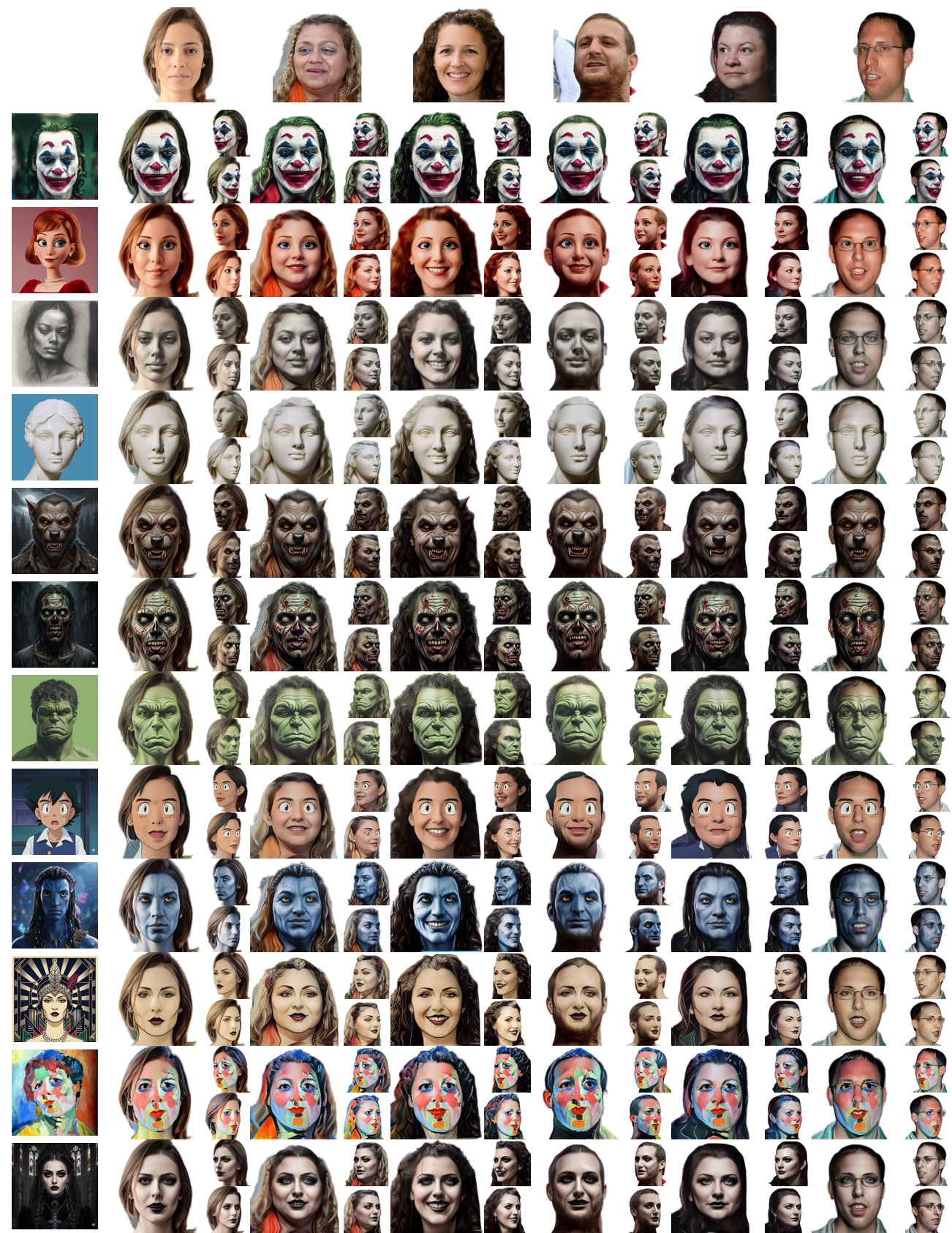}
      \vspace{-0.7cm}
    \caption{Additional qualitative results across a wide range of identities and style domains.}
    \label{fig:additional-1}
\end{figure*}

\begin{figure*}[t!]
    \centering
    \includegraphics[width=1.0\linewidth]{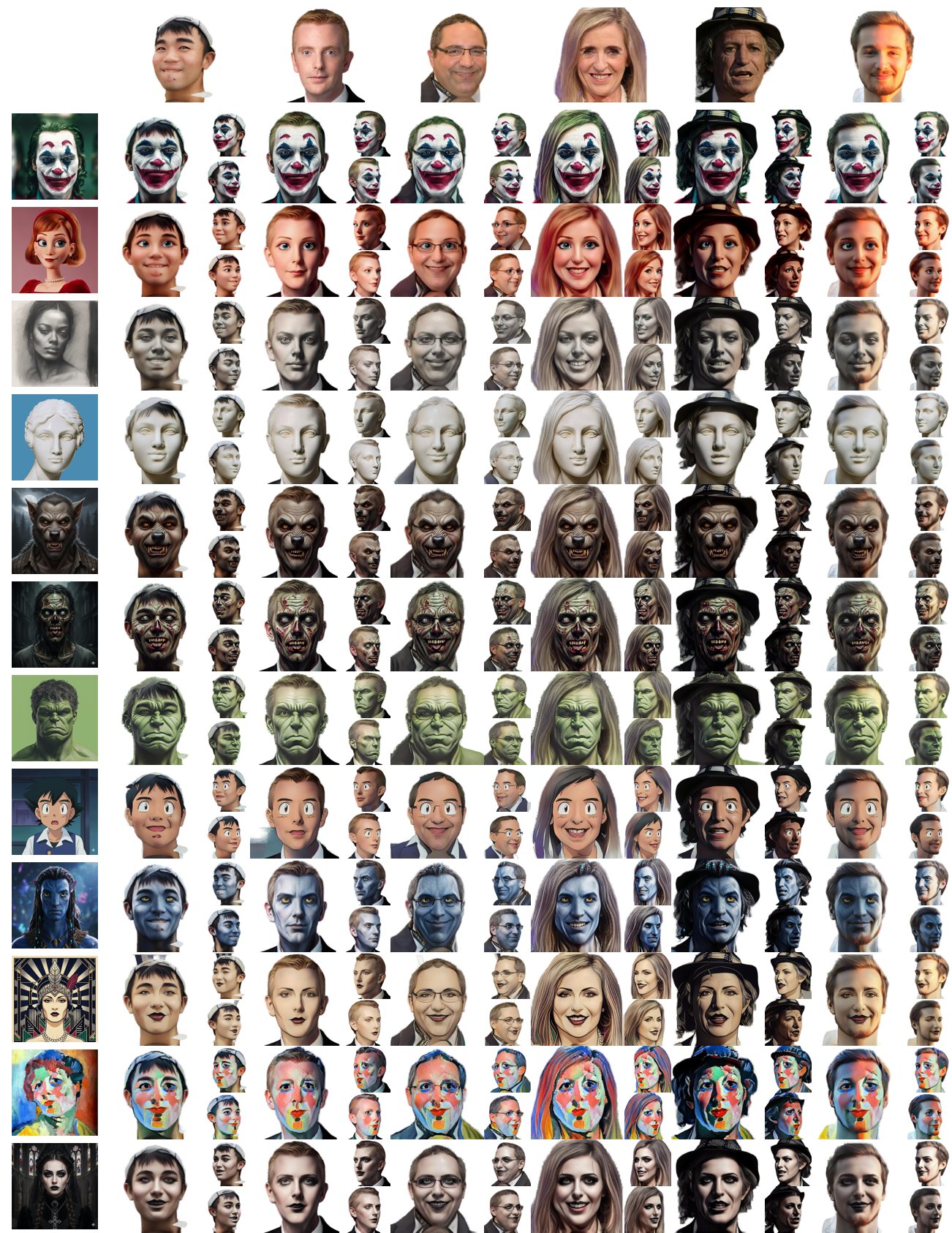}
      \vspace{-0.7cm}
    \caption{Additional qualitative results across a wide range of identities and style domains.}
    \label{fig:additional-2}
\end{figure*}

\cref{fig:additional-1,fig:additional-2} present the style reference images for a diverse set of artistic domains and the corresponding outputs produced by our model for multiple input identities.

\end{document}